\documentclass{article}

\PassOptionsToPackage{numbers,sort&compress}{natbib}
\usepackage[dblblindworkshop, final]{neurips_2026}
\workshoptitle{TAE (Trust-AI-Eval): Can We Trust AI Evaluation?}

\usepackage[utf8]{inputenc}
\usepackage[T1]{fontenc}
\usepackage[hidelinks]{hyperref}
\usepackage{url}
\usepackage{booktabs}
\usepackage{longtable}
\usepackage{multirow}
\usepackage{amsfonts}
\usepackage{amsmath}
\usepackage{amssymb}
\usepackage{microtype}
\usepackage{xcolor}
\definecolor{crblue}{HTML}{0057B8}
\newif\ifcrreview
\crreviewfalse
\newcommand{\crcolor}{\ifcrreview\color{crblue}\fi}
\newcommand{\crrev}[1]{{\ifcrreview\color{crblue}\fi #1}}
\newenvironment{crrevision}{\begingroup\ifcrreview\color{crblue}\fi}{\endgroup}

\usepackage{graphicx}
\usepackage[nameinlink]{cleveref}

\usepackage{todonotes}

\crefname{figure}{Fig.}{Figs.}
\Crefname{figure}{Fig.}{Figs.}
\crefname{table}{Table}{Tables}
\Crefname{table}{Table}{Tables}
\crefname{section}{Sec.}{Secs.}
\Crefname{section}{Sec.}{Secs.}
\crefname{equation}{Eq.}{Eqs.}
\Crefname{equation}{Eq.}{Eqs.}

\newcommand{\ELC}{\texttt{ELC}}
\newcommand{\REP}{\texttt{REP}}

\newif\ifsubmissionfinal
\submissionfinaltrue

\let\origcitep\citep
\renewcommand{\citep}[1]{\textcolor{black}{\origcitep{#1}}}
\let\origcitet\citet
\renewcommand{\citet}[1]{\textcolor{black}{\origcitet{#1}}}

\title{\crrev{Metamorphic testing of learned AC optimal power flow surrogates reveals representation dependence}}
\hypersetup{pdftitle={Metamorphic testing of learned AC optimal power flow surrogates reveals representation dependence}}
\author{%
  Stephen Peng\thanks{Equal contribution.}\quad Vineet Jagadeesan Nair\footnotemark[1] \\
  AI Science Institute, Boston Consulting Group \\
  \texttt{Peng.Stephen@bcg.com}, \texttt{JagadeesanNair.Vineet@bcg.com}
}

\begin{document}
\maketitle

\begin{abstract}
\begin{crrevision}
Evaluating a learned AC optimal power flow (AC-OPF) surrogate on one grid serialization cannot show whether its accuracy depends on that encoding. We test this dependence with thirteen metamorphic relations, each specifying when a transformation is valid and how to compare its outputs. Our audit covers 17 model and pipeline configurations: two released trained checkpoints and 15 random-weight architecture or mechanism controls. Every configuration violates at least six relations, even at reporting tolerances ten times tighter or looser. For \texttt{GridSFM-Open v1.1}, source inspection and intermediate-tensor interventions point to an indexing bug and three noncanonical or numerically unstable preprocessing choices; a repaired checkpoint remains untested. In a separate case500 training experiment, augmentation lowers transformed-view error in three of four models and median discrepancy and constraint-violation P95 in all four. These benefits depend on the training protocol: some relations worsen, and every evaluated prediction remains AC infeasible. A GPU/CPU eigensolver mismatch accounts for most of the large raw-Laplacian-baseline gain. We recommend reporting representation consistency alongside accuracy, feasibility, and cost. These results establish neither deployment readiness nor universal invariance.
\end{crrevision}
\end{abstract}

\section{Introduction}

AC-OPF chooses generator dispatch and network states to minimize cost while satisfying physical equations and operating limits. Learned surrogates approximate this calculation.
\crrev{Published evaluations report accuracy, feasibility, and speed on held-out scenarios, each stored in one serialization \citep{pilotoetal2024canos,liuetal2023topologyaware,arowolocremer2026generalization}.}
\crrev{A bus can move to another row without changing the grid. Eligible branches can be reversed, and each island can use a different angle reference. Component order and model-visible base-voltage metadata can also change without changing the declared problem, under the conditions specified below. Once predictions are returned to their physical devices and a common angle reference, a representation-consistent deterministic surrogate should give the same answer.}
\crrev{Testing one encoding leaves this property unmeasured. Error on one serialization alone cannot tell whether accuracy reflects the physical problem or a convention used to encode it. This gap between the measured quantity and the intended claim is a construct-validity problem in the evaluation protocol.}

\crrev{Metamorphic testing checks necessary relationships between outputs for linked inputs without requiring a unique reference output \citep{chenetal2018metamorphic,seguraetal2016survey}.} We define thirteen AC-OPF transformations in two tiers. Representation relations re-encode the same grid. Electrical-equivalence relations change component or graph structure while preserving the aggregate electrical problem. Each relation specifies an output map that restores the original identities and angle reference, aggregating predictions where needed. \crrev{Algebraic conditions establish when the relation is valid, and sampled solver controls check its implementation (\cref{app:audit-controls} gives the archived admission rules and coverage limits). One admitted violation refutes invariance for that pipeline and relation. A finite set of passes shows only that the tests found no violation.}

Our main contributions are:
\begin{itemize}
  \item \crrev{We define thirteen metamorphic relations with explicit output maps, numerical admission criteria, and evidence tiers that separate open-weight checkpoints, random initializations, and reimplementations of common AC-OPF neural architecture components (\cref{sec:method,app:catalog}).}
  \item \crrev{We audit 17 model and pipeline configurations, including two open-weight checkpoints. Every configuration fails at least six relations, and for both released checkpoints the trained and random-weight versions agree on all 13 aggregate threshold classifications (\cref{sec:screen,sec:trained-random}).}
  \item \crrev{We localize the first detected discrepancies in the grid foundation model \texttt{GridSFM-Open v1.1} to derived preprocessing features, implicating one indexing bug and three noncanonical or numerically unstable feature choices; we have not validated a repaired checkpoint (\cref{app:localization}).}
  \item \crrev{We test the relations as training augmentations under an AC constraint-penalty objective. Transformed-view error falls in three of four models, and median discrepancy and AC constraint-violation P95 fall for all four. Every prediction remains AC infeasible, and an eigensolver mismatch accounts for most of the largest gain (\cref{sec:aug-results}).}
  \item \crrev{We propose reporting practices and a signed-manifest scheme through which closed-weight groups can publish auditable consistency evidence without exposing parameters (\cref{sec:discussion}).}
\end{itemize}
\crrev{Code is available in \href{https://anonymous.4open.science/r/metamorphic-testing-ac-opf-paper-D2E5/README.md}{an anonymous repository}.}

\section{Related work}
\label{sec:related-work}

\subsection{{Learned AC-OPF surrogates}}
\label{app:related-work-surrogates}

Learned AC-OPF models map demands, generator limits, costs, constraints, and network parameters to voltage magnitudes, angles, and dispatch. They may replace a nonlinear optimizer, provide its starting point, or work with a correction step. Fast predictions with low supervised error can still violate power balance or operating limits and can have high dispatch cost. Physics-informed losses and differentiable completion or correction layers address some of these problems \citep{nellikkathchatzivasileiadis2022physicsinformed,dontietal2021dc3}.

Early surrogates often used fully connected networks tied to a fixed topology and variable order. Graph neural networks use bus--branch connectivity and \crrev{shared message-passing rules, allowing permutation-equivariant computation} \citep{battagliaetal2018relational,maronetal2018invariant}. Homogeneous and heterogeneous AC-OPF models use this structure \citep{owerkoetal2020optimal,liuetal2023topologyaware,ghamizietal2024opfhgnn,pilotoetal2024canos}. Distinguishing buses, generators, loads, shunts, lines, and transformers preserves useful physical information, but also requires consistent handling of each family's order, identity, and aggregation.

\crrev{Local message passing can struggle to represent long-range electrical coupling. Global attention, graph-transformer blocks, and positional encodings extend its reach} \citep{rampaseketal2022recipea,wenetal2026heterogeneous,arowolocremer2026generalization}. Laplacian eigenvectors, effective resistance, random-walk statistics, and landmark summaries supply global information, but some also depend on arbitrary node indices, eigenvector signs or bases, selected reference nodes, or row order. These features can break equivariance in the full feature-to-prediction pipeline even when the graph layers themselves are equivariant.

\subsection{{Metamorphic testing and equivariance}}
\label{app:related-work-metamorphic}

Metamorphic testing addresses the test-oracle problem: how to test software when the correct output is unavailable or costly to obtain. It checks necessary relationships between outputs for linked inputs \citep{chenetal2018metamorphic,seguraetal2016survey}. If an input transformation has a known effect on a correct output, a violated relationship can reveal a defect without a unique reference label. For example, symmetry-based relations have been used to test human-trajectory prediction models \citep{spiekeretal2024trajectory}.

\crrev{The AC-OPF relations have different mathematical forms.} A bus permutation requires the same permutation of bus outputs; after mapping them back to physical identities, predictions should be unchanged. \crrev{A reference-angle change is a gauge transformation: it shifts absolute angles while preserving angle differences, flows, and dispatch after canonicalization.} Component and parallel-branch splits can change output dimension and require many-to-one aggregation. \crrev{We reserve \emph{built-in equivariance} for a property that holds at every parameter setting. Consistency observed in one fitted checkpoint is an empirical result that depends on the tested input distribution.}

\crrev{Learned-OPF studies typically report} accuracy, optimality, feasibility, speed, and generalization to changed loads or contingencies \citep{pilotoetal2024canos,liuetal2023topologyaware,arowolocremer2026generalization}. \crrev{An outage changes the physical optimization problem; an admitted relation in our audit preserves it after a known output map. In the literature we reviewed, we found no AC-OPF study combining a relation catalog with solver controls, trained-checkpoint tests, architecture-level failure localization, and an augmentation experiment. Our contribution is this OPF-specific construction and audit, building on established metamorphic-testing and graph-equivariance methods.}

\begin{crrevision}
\paragraph{Relation to formal guarantees and invariant spectral features.}
Worst-case verification asks whether a predictor meets a constraint or quality bound throughout an input domain; \citet{venzke2020worstcase}, for example, study neural DC-OPF guarantees. Our tests find counterexamples on sampled, conditionally equivalent AC-OPF encodings and cannot certify a whole domain. The raw Laplacian coordinates in our diagnostic baseline have known sign and eigenspace-basis ambiguities. SignNet and BasisNet address these symmetries explicitly \citep{limetal2023sign}; RWSE avoids these ambiguities through a different construction. We do not benchmark trained SignNet or BasisNet, so our LapPE findings should not be generalized to all spectral graph models.

\paragraph{Related scientific-surrogate metamorphic testing.}
\citet{li2026domainvalidity} use metamorphic tests for scientific ML surrogates with explicit applicability conditions and numerical thresholds. We share their requirement to check that a relation applies before interpreting a violation. Here we develop the AC-OPF relations and output maps, distinguish trained-checkpoint tests from random-weight controls, and audit complete grid-model pipelines. Validity-gated metamorphic testing and its use beyond power systems precede this work; our experiments cover power systems only.
\end{crrevision}

\subsection{Evaluation validity beyond power systems}
\label{sec:eval-validity}

\crrev{Evaluation choices can affect conclusions in other domains as well. Meaning-preserving prompt-format changes} can shift measured accuracy by tens of points \citep{sclar2024quantifying}; model rankings from one prompt may not hold across prompts \citep{mizrahietal2024state}; and undetected data contamination can inflate scores \citep{sainzetal2023nlp}. Systematic reviews also find that many benchmarks leave construct validity and reporting practices unexamined \citep{reueletal2024betterbench,erikssonetal2025trust}.

A heuristic perturbation may change the task it is meant to test. Our relations preserve the declared AC-OPF problem under explicit conditions, and \crrev{sampled numerical controls check their implementation before we interpret the model comparisons. AC-OPF supplies known physical invariances that can be checked numerically.} The same audit structure could test consistency in other domains where task-preserving transformations can be verified.

\section{Metamorphic testing for AC-OPF}
\label{sec:method}

\begin{crrevision}
For a deterministic surrogate $f$, relation $r$ transforms serialized case $d$ into an equivalent input $T_r d$. Its output map $P_r$ returns predictions to their physical devices, aligns angles independently within each connected island, and aggregates split-device outputs where needed. We measure departure from $P_r f(T_r d)=f(d)$ by
\begin{equation}
\Delta_r(d)=\max_{c\in\mathcal C}
\frac{\operatorname{MAE}_c\!\left(P_r f(T_r d),f(d)\right)}{\tau_c},
\qquad \mathcal C=\{V_m,V_a,P_g,Q_g\}.
\label{eq:empirical-discrepancy}
\end{equation}
Here $V_m,V_a$ denote voltage magnitude and angle, and $P_g,Q_g$ denote active and reactive generation. Each channel's mean absolute error (MAE) averages absolute differences over its elements. We set $\tau_c=10^{-6}$ in native units: per unit for $V_m,P_g,Q_g$ and radians for $V_a$. Angles are re-anchored without further modulo wrapping. A value of $\Delta_r=1$ reaches the reporting threshold; it is neither a percentage error nor an AC-feasibility limit.

\textbf{Verdicts and summaries.} A cell in the 17-configuration screen passes only when every executed, applicable draw satisfies $\Delta_r\leq1$. We record partial coverage where controls are unavailable and check maximum component errors separately at $10^{-5}$. The five-grid trained-versus-random comparison summarizes magnitude differently: it averages draws within each source scenario, takes the median over sources, and then takes the worst grid, keeping weight seeds separate. A below-threshold aggregate does not mean that every draw passed. \Cref{app:audit-controls} specifies the numerical controls, and \cref{app:screen-sensitivity} reports threshold sensitivity and component tails.

\textbf{Validity and scope.} Eight representation-equivalence (\REP) relations change labels, orientation, angle-reference conventions, or a field used by a particular pipeline. Five electrical-equivalence (\ELC) relations change component or graph bookkeeping while preserving the declared aggregate problem. Algebraic conditions establish applicability; the saved solver controls check implementation on a sample of evaluation views. \Cref{app:catalog} specifies the transformations, applicability conditions, and output maps. \Cref{app:acopf-nonuniqueness,app:data-fibers} discuss non-uniqueness and the map from serialized data to the physical problem.

\textbf{Worked output map example.} If \texttt{bus-relabel} stores old buses as $(3,1,2)$, restore voltage predictions as $(\widehat V'_2,\widehat V'_3,\widehat V'_1)$ and match generator outputs by physical device identity. If the reference bus also changes, subtract the restored reference angle from every angle in that island before comparing. A discrepancy measures deterministic representation dependence. Either prediction could still be feasible, because a nonconvex AC-OPF problem can admit several physically valid solutions.

\begin{table}[htbp]
\crcolor
\caption{The thirteen relations at a glance. The two tiers describe the output-map contract; neither tier asserts that every relation is a bijective group action.}
\label{tab:catalog-main}
\centering\small
\begin{tabular}{@{}p{0.15\linewidth}p{0.39\linewidth}p{0.38\linewidth}@{}}
\toprule
Tier & Relations & What is held fixed / restored \\
\midrule
\REP & \texttt{component-relabel}, \texttt{bus-relabel}, \texttt{row-order} & Physical identities; inverse permutations \\
\REP & \texttt{branch-reversal}, \texttt{tap-side-flip} & Branch two-port equations; endpoint/attribute transport \\
\REP & \texttt{ref-bus-move}, \texttt{island-ref-mix} & Per-island angle gauge after mapping back \\
\REP & \texttt{basekv-scale} & Exported per-unit target; conditional on schema semantics \\
\ELC & \texttt{component-split}, \texttt{parallel-split} & Aggregate injection, costs or admittance; sum split outputs \\
\ELC & \texttt{load-as-gen}, \texttt{null-pad} & Net bus injection; undo device recoding/padding \\
\ELC & \texttt{island-union} & Original disconnected subproblem; restrict to its outputs \\
\bottomrule
\end{tabular}
\end{table}

A relation tests the full adapter, preprocessing, prediction, and output-restoration path. For example, \texttt{basekv-scale} is valid here only because the exported target equations are already in per unit and the changed field is model-visible metadata; the relation does not extend to rescaling physical voltages. We check conditional applicability before interpreting a discrepancy. A violation outside a model's documented input contract diagnoses interface compatibility and may not indicate a defect in its intended use.
\end{crrevision}

We use random initialization as a counterexample screen for built-in consistency. One violation refutes architectural invariance to that relation, \crrev{whereas a pass means only that the screen found no violation.}
\crrev{The DC3 controls, for example, test only the partition-and-completion choice with a stand-in; they do not run DC3's Newton power-flow completion or correction loop.}
We separately compare released (open-weight) and random-weight versions of \texttt{GridSFM-Open v1.1} and \texttt{LUMINA-2M v0.1.3} \crrev{to test whether these training states change the aggregate threshold classification} \citep{yangetal2026gridsfm,microsoftgridsfmopen2026modelcard,microsoftgridsfm2026software,lietal2026lumina,argonnelumina2m2026modelcard}.
Scenarios come from the N-1 topology-perturbed \texttt{OPFData} release, with \texttt{PGLib} networks and \texttt{PowerModels.jl} used for numerical solver controls \citep{lovettetal2024opfdata,babaeinejadsarookolaeeetal2021power,coffrinetal2018powermodels}. The full model inventory, sampling scheme, and solver admission rules appear in \cref{app:model-inventory,app:audit-controls}.

\crrev{Numerical convergence alone does not admit a relation. All 162 pairs of solves in the expanded-relation controls converge, but only 17 of 18 grid--relation pairs meet the admission criteria. Case14 \texttt{component-split} is withheld because its relative objective difference, $7.26\times10^{-9}$, exceeds the $10^{-9}$ limit. The screen reuses these sampled controls without solving every model draw; fourteen \texttt{component-split} cells therefore retain only case500 evidence.}

\section{Audit results}
\label{sec:results}

\subsection{Built-in consistency screen}
\label{sec:screen}

\begin{figure}[htbp]
\centering
\includegraphics[width=0.9\linewidth]{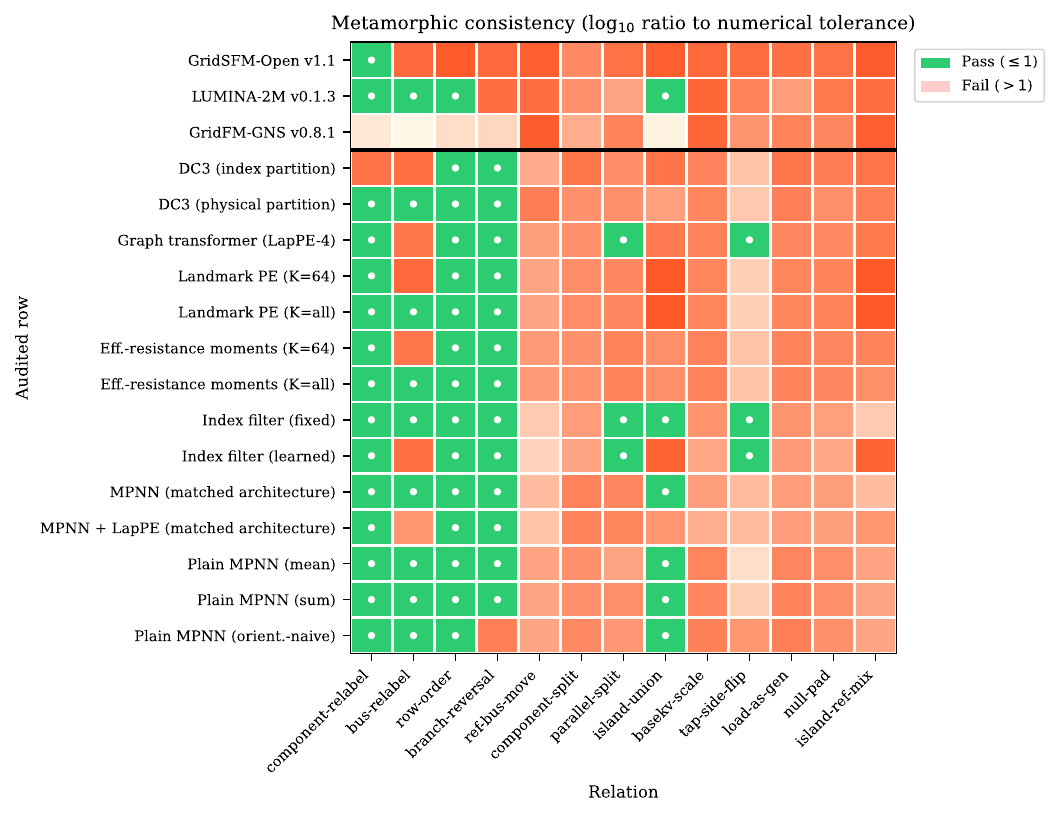}
\caption{{Metamorphic consistency for 17 model and pipeline configurations (two trained checkpoints and 15 random-weight architecture or mechanism controls) across 13 relations on case14/500. Each case uses 3 source scenarios, 3 sampled transformations per admitted case--relation pair, and 3 random initializations where applicable. Green denotes a pass at the prespecified $10^{-6}$ tolerance.}}
\label{fig:screen-heatmap}
\end{figure}

\cref{fig:screen-heatmap} summarizes 17 model and pipeline configurations\crrev{, of which two are trained open-weight checkpoints and 15 are random-weight architecture or mechanism controls (\cref{tab:screen-models}).} No row passes all thirteen relations, and every row fails at least six. The six shared failures are \texttt{basekv-\allowbreak scale}, \texttt{ref-\allowbreak bus-\allowbreak move}, \texttt{island-\allowbreak ref-\allowbreak mix}, \texttt{component-\allowbreak split}, \texttt{load-\allowbreak as-\allowbreak gen}, and \texttt{null-\allowbreak pad}. \crrev{At a relaxed native-channel threshold of $10^{-3}$, all 17 configurations still fail at least four relations; excluding the pipeline-specific \texttt{basekv-scale} relation leaves at least three failures per configuration. At the primary $10^{-6}$ threshold, the same exclusion leaves 140 of 204 cells failed and at least five failed relations per configuration. This reaggregation of saved results tests threshold sensitivity only and does not calibrate numerical noise (\cref{app:screen-sensitivity}).}
\Cref{app:failure-patterns} traces the pattern to specific design mechanisms.

\subsection{Architecture-level failure patterns}
\label{app:failure-patterns}

\crrev{Relabeling gives the clearest distinction among designs. Those that tie features to physical components rather than array position pass \texttt{component-\allowbreak relabel}, \texttt{bus-\allowbreak relabel}, and \texttt{row-\allowbreak order}. The \texttt{row-\allowbreak order} relation fails only in the GridSFM-Open and GridFM processing paths, which use row position directly.}

\crrev{The \texttt{bus-\allowbreak relabel} relation fails in 8 of 17 rows. Each failing row uses index-derived or spectral coordinates through Laplacian or landmark positional encodings, index-partitioned heads, or index filters. The matched MPNN and LapPE+MPNN rows isolate one such choice: adding raw Laplacian coordinates turns \texttt{bus-\allowbreak relabel} and \texttt{island-\allowbreak union} from passes into failures, because eigenvector signs and eigenspace bases depend on ordering. Index-selected landmarks are buses chosen by array position, so relabeling changes which physical buses are selected. Using all buses removes this selection dependence in the matched control; we have not validated a repaired trained checkpoint.}

\crrev{The \texttt{island-\allowbreak union} relation distinguishes local from global computation. Models with purely local, permutation-safe aggregation are unaffected by adding a disconnected island, while rows with global positional summaries or spectral coordinates fail. The \texttt{branch-\allowbreak reversal} relation fails only in rows that consume directed edge features without symmetrizing them. LUMINA shows the same pattern in the trained/random comparison (\cref{sec:trained-random}).}

\crrev{The six relations that fail in every configuration expose two kinds of preprocessing dependence. First, \texttt{ref-\allowbreak bus-\allowbreak move} and \texttt{island-\allowbreak ref-\allowbreak mix} change the angle reference, while \texttt{basekv-\allowbreak scale} changes model-visible metadata that does not affect the standard per-unit target after \texttt{OPFData} export. These failures show that the audited pipelines do not consistently canonicalize angle references or discard target-irrelevant fields. Second, \texttt{component-\allowbreak split}, \texttt{load-\allowbreak as-\allowbreak gen}, and \texttt{null-\allowbreak pad} change how device records represent a fixed bus-level injection. Their failures show dependence on component count or type beyond the electrical aggregates in bus balance. Standardizing inputs may address the first kind. The second may require a common representation for electrically equivalent components and aggregation of their contributions at each bus (\cref{app:exact-construction}).}

\subsection{Trained versus random-weight consistency}
\label{sec:trained-random}

\crrev{This comparison uses three source scenarios from each of case14, case118, case500, case2000, and case4661, with three transformation draws per source--relation pair: $3\times13\times3=117$ comparisons per grid and model/seed run. Each released checkpoint and its three random initializations receive the same scenarios and draws. Trained runs stored under three seed labels repeat the same checkpoint and draws; the independent units remain three sources per grid. The grids differ in their relation to each checkpoint's documented training coverage, so this is not a common unseen-topology benchmark (\cref{app:audit-controls}).}

\crrev{Trained and random-weight versions of both open-weight backbones agree on all 13 threshold classifications of the source-median, worst-grid aggregate} (\cref{fig:weight-study-verdict}). \crrev{The aggregate is below threshold for GridSFM only on} \texttt{component-\allowbreak relabel}; \crrev{for LUMINA it is additionally below threshold on} \texttt{bus-\allowbreak relabel}, \texttt{row-\allowbreak order}, and \texttt{island-\allowbreak union}. \crrev{These below-threshold aggregates agree with the finite permutation and aggregation checks and do not establish an all-input guarantee. Both trained and random LUMINA variants fail \texttt{branch-reversal}, consistent with a directed message-passing path that does not transport the branch attributes needed to compensate for endpoint exchange. Passing label permutations therefore does not imply consistency under every \REP{} relation.}

\crrev{Training changes most discrepancy magnitudes without consistently reducing them.} We exclude \texttt{component-\allowbreak relabel} from ratio comparisons because its trained and random discrepancies are all below $10^{-6}$, making ratios between near-zero values uninformative. Among the remaining finite case--relation cells, trained discrepancy exceeds the random control in 55 of 60 GridSFM cells and 50 of 58 LUMINA cells. \crrev{The median ratios are $7.50\times$ and $6.23\times$, respectively, with substantial variation across grids. Several LUMINA reference-related cells have ratios below one.}

A large ratio can result from a near-zero random denominator, while a ratio below one can still leave a discrepancy far above tolerance. \crrev{Both absolute and relative values are therefore needed.} Random networks are inaccurate and can appear consistent because their outputs are nearly constant; these ratios do not compare model quality. For the tested checkpoints, \crrev{training changed magnitudes but eliminated no above-threshold aggregate discrepancy. \Cref{app:trained-random} reports per-relation values in both views, and \cref{fig:weight-study-ratio} shows how training changes discrepancies on each grid.}

\subsection{Localization in released pipelines}
\label{app:localization}

\crrev{For GridSFM, the bus-to-bus edge index is permutation-equivariant, and shuffling edge rows after feature construction leaves network predictions unchanged.}
\crrev{The first detected discrepancy appears in derived features before the prediction network. Four candidate mechanisms explain where it can enter: effective-resistance summaries select up to 64 landmarks partly by bus index; the combined line/transformer offsets count appended arrays rather than edges; applying \texttt{sign} to a near-zero DC flow turns numerical noise into a categorical feature; and the cycle basis uses a spanning tree weighted by edge-row order.} We reproduced these observations against pinned commit \texttt{1ca775f} \citep{microsoftgridsfm2026software}.

\begin{crrevision}
We diagnose these mechanisms by substituting aligned intermediate tensors; this procedure does not produce a repaired-and-retrained checkpoint. For case500 source 9245 under bus relabeling (draw 0), the saved cumulative intervention changes the legacy equal-channel discrepancy score from $30.4769$ unpatched to $30.4712$ after node/cycle feature substitution and to $5.2535\times10^{-5}$ after cycle-incidence substitution. This score averages channel errors divided by $10^{-3}$, unlike the paper's maximum-channel $\Delta_r$ at $10^{-6}$. The final-to-initial ratio, $1.724\times10^{-6}$, measures the relative residual and is not an absolute pass threshold.

Other relations retain substantial residuals under the same interventions. In a separate network-only probe, edge-row shuffling gives zero movement, while branch-node relabeling gives a nonzero $3.51\times10^{-5}$ discrepancy on the legacy scale. These tests locate substantial sensitivity before the prediction network. They neither prove that the entire network is invariant nor validate four independent repairs.
\end{crrevision}

\crrev{The candidate mechanisms suggest four repairs: select landmarks independently of arbitrary indices or summarize all buses; track cumulative edge counts when joining component families; apply a physically justified dead band or smooth mapping to near-zero flows; and construct cycle information independently of row order. A patch-and-retest experiment is still needed to establish whether these changes restore consistency. A checkpoint may have adapted to its existing features, so each repair should be followed by retraining or fine-tuning and an accuracy check.}

\crrev{LUMINA localizes differently from GridSFM. Its below-threshold results under component relabeling, bus relabeling, row ordering, and disconnected-island composition hold for both trained and random weights (\cref{sec:trained-random}), and its branch-reversal failure is consistent with orientation-sensitive directed processing.} The random-weight comparison and network-only tests thus separate preprocessing defects, architectural transport choices, and changes in magnitude acquired during training.

\crrev{Representation changes also alter the cost implied by GridSFM's active-power predictions. For the trained checkpoint, the median of the absolute cost change normalized by each scenario's solver reference cost is $23.4\%$ under \texttt{row-order}, $50.2\%$ under \texttt{island-union}, and $51.3\%$ under \texttt{island-ref-mix} (\cref{tab:gridsfm-cost-movement}). These predictions are AC infeasible: the values measure encoding sensitivity of implied cost, not realized dispatch losses or optimality gaps. \Cref{app:dispatch-cost} defines the normalization and reports all relations.}

\subsection{Relations as data augmentation}
\label{sec:aug-results}

\crrev{We test whether training on the relations reduces representation dependence.} We train baseline and augmented versions of \crrev{an MPNN,} an MPNN with raw Laplacian positional encodings (LapPE+MPNN) \citep{dwivedietal2023benchmarking,rampaseketal2022recipea}, an MPNN with random-walk structural encodings that avoid sign and eigenvector-basis ambiguities (RWSE+MPNN) \citep{dwivedietal2022graph,rampaseketal2022recipea}, and LUMINA. All models minimize equal-channel MSE against the solver solution plus an AC constraint-violation penalty. \crrev{For each training instance, the augmented variant applies one relation with probability $0.5$. It draws from a frozen architecture-specific set chosen from earlier trained-baseline measurements, independently of the random-weight screen.}

We evaluate 50 held-out scenarios per seed in \crrev{canonical and transformed forms} under all thirteen relations. \crrev{Accuracy is the unweighted mean of native-unit MAEs across $V_m,V_a,P_g,Q_g$. This convention combines p.u.\ and radians, so it should be read alongside the per-channel values in \cref{tab:augmentation-summary}. \Cref{app:augmentation-protocol} gives the training objective, relation sets, hyperparameters, and metrics.}

\begin{table}[htbp]
\crcolor
\caption{Aggregate augmentation outcomes under the AC constraint-penalty objective on held-out case500 scenarios (baseline $\to$ augmented; three seeds). \crrev{Median $\Delta_r$ summarizes each architecture's own trained relation set and supports within-model comparisons. }Constraint columns report mean $\pm$ sample standard deviation. Within each seed, we take each source scenario's worst violation over its 53 canonical and transformed views and then P95, the linearly interpolated 95th percentile, across 50 source scenarios. Every arm has 0\% AC-feasible source scenarios at $10^{-6}$. Equality is the larger of the maximum nodal active- and
reactive-power balance violations in \cref{eq:appendix-acopf-active-balance,eq:appendix-acopf-reactive-balance}, expressed in per unit. The pooled inequality metric is dimensionless: each family is divided by its $10^{-6}$ tolerance before taking the maximum, and displayed values are in millions of tolerances. Units, metric definitions, and full results appear in \cref{app:augmentation-protocol,tab:augmentation-summary,tab:aug-appendix}.}
\label{tab:augmentation-main}
\centering
\small
\setlength{\tabcolsep}{4.5pt}
\renewcommand{\arraystretch}{1.0}
\resizebox{\linewidth}{!}{%
{
\begin{tabular}{@{}lccccc@{}}
\toprule
Model & \shortstack{Canonical\\MAE} & \shortstack{Transformed\\MAE} & \shortstack{Median\\$\Delta_r$} & \shortstack{Eq. violation P95\\(p.u.)} & \shortstack{Ineq. violation P95\\($10^6\!\times$ tol.)} \\
\midrule
MPNN & $0.01295 \to 0.01345$ & $0.01312 \to 0.01353$ & $852 \to 595$ & $58.1\!\pm\!18.7 \to 42.2\!\pm\!2.3$ & $19.3\!\pm\!4.2 \to 11.9\!\pm\!1.6$ \\
LapPE+MPNN & $0.09983 \to 0.01309$ & $0.10211 \to 0.01318$ & $1{,}000 \to 805$ & $113.9\!\pm\!42.1 \to 39.5\!\pm\!5.2$ & $42.1\!\pm\!4.6 \to 12.7\!\pm\!3.6$ \\
RWSE+MPNN & $0.01255 \to 0.01250$ & $0.01271 \to 0.01257$ & $883 \to 563$ & $51.8\!\pm\!10.9 \to 33.5\!\pm\!2.4$ & $18.8\!\pm\!2.8 \to 10.2\!\pm\!0.8$ \\
LUMINA & $0.01834 \to 0.01889$ & $0.03594 \to 0.02089$ & $4{,}630 \to 1{,}260$ & $871\!\pm\!342 \to 164\!\pm\!11$ & $276\!\pm\!91 \to 20.1\!\pm\!2.1$ \\
\bottomrule
\end{tabular}
}}
\end{table}

\crrev{Augmentation lowers mean MAE on transformed scenarios by $87.0\%$ for LapPE+MPNN, $41.9\%$ for LUMINA, and $1.1\%$ for RWSE+MPNN, and raises it by $3.2\%$ for MPNN (\cref{tab:augmentation-main}).} \crrev{Canonical MAE improves by $0.4\%$ for RWSE+MPNN and worsens by $3.9\%$ for MPNN and $3.0\%$ for LUMINA.}
\crrev{Median normalized discrepancy $\Delta_r$ falls for all four models. Some individual relations worsen, out-of-distribution grid generalization is mixed, and no prediction from our runs satisfies both the inequality and equality tolerances.} Per-seed and relation-level effects appear in \cref{app:augmentation-accuracy}, with feasibility and out-of-distribution results in \cref{app:augmentation-feasibility}.

\crrev{The relation-level results show what the aggregate obscures. Each arm's per-relation summary takes the largest source median across seeds. LapPE+MPNN worsens on \texttt{null-pad}, \texttt{component-split}, and \texttt{load-as-gen}; RWSE+MPNN worsens on \texttt{tap-side-flip}, \texttt{ref-bus-move}, and \texttt{island-ref-mix}. For example, their discrepancies increase by $29.4\%$ on \texttt{null-pad} and $21.2\%$ on \texttt{tap-side-flip}, respectively. In these per-relation summaries, MPNN improves on all seven relations used in its training and LUMINA on all nine (\cref{tab:aug-appendix}). A lower median therefore does not establish improvement on every relation.}

\begin{crrevision}
\crrev{Most of LapPE+MPNN's large improvement in \cref{tab:augmentation-main} comes from a mismatch between the GPU eigensolver used in training and the CPU eigensolver used in this audit, which may also inflate its baseline violation P95 values (\cref{app:cohort-provenance}).} Both eigensolver paths leave eigenvector signs and bases unfixed. With only the backend changed, the same checkpoints' clean-cohort baseline MAE rises from $0.0125$ to $0.0885$, while the augmented arm changes by $3\%$. Median augmented discrepancies remain $563$ to $1{,}260$ times the reporting threshold. Neither these reductions nor the lower constraint P95 values establish exact consistency, AC feasibility, or deployment readiness.
\end{crrevision}

\crrev{Scenario-clustered constraint-violation P95 also falls under augmentation for all four models under the same loss function. We have not tested whether training on invariance-preserving transformations causes this reduction, and the reduction does not show improved physical understanding.}

\section{Discussion}
\label{sec:discussion}

\crrev{Representation consistency asks whether a deterministic pipeline returns the same physically matched prediction for equivalent encodings. It is separate from accuracy, feasibility, and optimality. Aggregation keyed to physical components, permutation-safe features, symmetric handling of branch attributes, and a common angle reference can guarantee some relations by construction (\cref{app:exact-construction}). Augmentation instead encourages consistency on transformations represented in the training data.}

\crrev{The four augmentation models fail different relations. LUMINA's largest baseline failures are \texttt{branch-reversal} and \texttt{basekv-scale}; LapPE+MPNN's are \texttt{bus-relabel}, \texttt{island-union}, \texttt{island-ref-mix}, and \texttt{parallel-split}. A single-relation benchmark would therefore miss some models' largest failures.} Orbit-averaging theory motivates augmentation under compact group actions, but does not directly cover every conditional, pipeline-specific, or dimension-changing relation in this catalog \citep{chenetal2020grouptheoretic,elesedyzaidi2021provably,nordenforsetal2024optimization}\crrev{ (\cref{app:augmentation-theory})}.

\subsection{Toward benchmarks that test invariance}
\label{app:benchmark-protocol}

Current learned-OPF benchmarks generally sample physical scenarios but evaluate each under one serialization. We propose testing a small set of metamorphic variants as repeated views of each physical case. Developers can run this audit before release; operators and planners can use documented inference access. For each held-out scenario, check which relations apply, generate $K$ valid encodings, restore predictions to physical identities and a common angle reference, and inspect the worst discrepancy alongside per-channel accuracy, feasibility, optimality, and dispatch cost. \crrev{Keep failed admissions and infeasible predictions in the record. The source scenario remains the independent sampling unit: permutations and splits are repeated measurements, not extra data points.} \Cref{tab:reporting-protocol} lists the minimum reporting fields.

\crrev{Metamorphic variants serve two different roles. As an \emph{evaluation augmentation}, they test whether predictions and reported conclusions hold across valid encodings. As a \emph{training augmentation}, they change the training objective and may improve learned invariance. A finite pass after augmented training is evidence about the tested checkpoint and inputs; it certifies neither architectural equivariance nor operational use. A claim of built-in equivariance still needs an analytical argument. Benchmarks should retain random-initialization or architecture-level screens to search for counterexamples to that claim.}

\begin{table}[htbp]
\caption{{Minimum reproducibility fields for a metamorphic learned-OPF benchmark.}}
\label{tab:reporting-protocol}
  {%
\centering
\small
{
\begin{tabular}{@{}p{0.24\linewidth}p{0.69\linewidth}@{}}
\toprule
Record & Minimum contents \\
\midrule
Data and labels & Release, split, topology membership, scenario identifier, units, angle-reference convention, and label-generation pipeline. \\
Relation & Name, tier, parameters, applicability decision, transformed-case hash, and output map. \\
Solver control & Status, objective difference, equality and inequality constraint violations, and admission decision for both views. \\
Model & Code and checkpoint version, checksums, preprocessing contract, feature order, normalization, and output semantics. \\
Sampling & Source-scenario count, sampled transformations, weight seeds, and the clustering unit used for uncertainty. \\
Metrics & Per-channel median, tail, and worst discrepancy plus accuracy, AC constraint violations, bounds, feasibility, and cost. \\
Negative records & Unsupported relations, failed solver admission checks, missing joins, infeasible predictions, and threshold saturation retained rather than dropped. \\
Artifact & Case--relation--seed records, configurations, environment, and one command or script that regenerates tables and figures. \\
\bottomrule
\end{tabular}
}
}
\end{table}

\subsection{Open weights, transparency, and black-box auditing}
\label{app:open-weights}

\crrev{Source code and a training recipe let an auditor test an architecture at random initialization. Testing learned behavior also requires the trained state, exact preprocessing contract, normalization statistics, feature-column order, and a runnable path from a documented case file to canonical outputs. GridSFM-Open and LUMINA-2M supplied these resources, making the audit and localization findings possible} \citep{yangetal2026gridsfm,microsoftgridsfmopen2026modelcard,microsoftgridsfm2026software,lietal2026lumina,argonnelumina2m2026modelcard}.
\crrev{The releases let us compare trained and random versions of the same backbone, trace failures to input features or edge conventions, measure whether training reduced or amplified discrepancies, and evaluate augmentation without recreating an undocumented baseline. These findings, including the defects, depend on that openness.} \Cref{app:transparency-package} lists the minimum resources needed. Closed-weight groups can also participate through inference access: they can run the public transformations internally and publish signed or hashed manifests with applicability decisions, solver-control status, per-channel discrepancies, and failure tails. \crrev{A shared format would make these consistency reports comparable without disclosing proprietary parameters.}

\subsection{Limitations}
\label{app:limits}

\crrev{The thirteen relations cover the fibers enumerated in \cref{app:data-fibers}, but do not generate every equivalent AC-OPF encoding (\cref{app:fiber-coverage}). A pass means that this catalog found no violation in the tested cases, not that the pipeline is invariant to all equivalent inputs.}
\crrev{Only two topology-varying surrogates supplied both a compatible checkpoint and a runnable inference path, which limits the breadth of our trained-weight study.}
Their agreement with the matched random-weight controls \crrev{on all thirteen aggregate threshold classifications shows that these particular checkpoints did not eliminate the detected failures; it does not show that ordinary training can never learn a missing relation.}

\crrev{The augmentation study covers one training grid family, three seeds, one compute budget, and model-specific eligible relation sets. Some relations worsen, transfer to unseen grid sizes is mixed, and no arm produces an AC-feasible prediction. These results do not establish deployment readiness. PowerModels.jl and IPOPT provide high-accuracy numerical controls, not global-optimality certificates for the nonconvex problem. Distinct local optima can differ in weakly identified channels such as reactive power dispatch.}
A metamorphic discrepancy therefore establishes deterministic representation dependence of the surrogate without showing that either output is physically invalid.

\subsection{{Future work}}
\label{app:future-work}

Future benchmark versions can add transformations such as unequal component splits, inactive-bound changes, cost rescalings that preserve the optimal dispatch, or schema-level bus--branch conversions.
\crrev{Further work can determine which relation families admit generators or analytical guarantees, compare finite catalogs with property-based sampling over continuous fibers, and develop tolerance rules tied to solver conditioning. Model work should compare augmentation with mechanisms that enforce invariance by construction. Candidates include canonical unit and angle-reference preprocessing, frame averaging or learned canonicalization applied to the serialization group \citep{punyetal2022frame,kabaetal2023equivariance}, sign- and basis-invariant or provably stable spectral features in place of raw Laplacian coordinates \citep{limetal2023sign,huangetal2024stability}, output heads keyed by physical role and component identity rather than array position \citep{ghamizietal2024opfhgnn}, and permutation-invariant set aggregation over co-located devices \citep{zaheeretal2017deep}.}
\crrev{Feasibility layers and correction methods should be tested jointly with these representation fixes: post-processing cannot be assumed to preserve consistency.}
Larger training studies should add grid families, compute budgets, and checkpoint initializations while retaining negative relation-level outcomes.

\section{Conclusion}
\label{sec:conclusion}

\crrev{A single-serialization evaluation cannot separate a learned AC-OPF surrogate's accuracy on the physical problem from its dependence on the encoding. Our thirteen transformations test that dependence with explicit output maps, validity conditions, and sampled solver admission checks. Each of the 17 audited configurations violates at least six relations. Six fail in every configuration: two change angle-reference conventions, one changes a model-visible base-voltage field outside the per-unit target, and three change how device records represent a fixed bus injection.} For both released checkpoints, training changes discrepancy magnitudes, usually upward, \crrev{without changing the reported aggregate threshold classifications. These finite observations do not establish which relations training can or cannot learn. In \texttt{GridSFM-Open v1.1}, intermediate-tensor interventions point to one indexing bug and three noncanonical or numerically unstable preprocessing choices. A small network-only residual remains, and a repaired checkpoint is still untested.}

\crrev{Training with these transformations lowers transformed-view error in three of four models and median discrepancy and AC constraint-violation P95 in all four. Individual relations and cross-grid transfer show mixed results, an eigensolver mismatch accounts for most of the largest gain, and every evaluated prediction remains AC infeasible. Representation consistency should therefore be reported alongside accuracy, optimality, and feasibility. Once a relation is admitted, measuring consistency requires model inference on transformed inputs. Learned-OPF benchmarks should evaluate several serializations of each physical scenario and compare predictions after restoring their physical identities and angle reference. The same audit structure could apply wherever task-preserving transformations can be checked independently of the model; a second domain remains untested.}

\begin{ack}
\crrev{The authors received no funding for this work and declare no competing interests.}
\end{ack}

\clearpage
\bibliographystyle{unsrtnat}
\bibliography{references_zotero}

\clearpage
\appendix

\section{Relation catalog and output maps}
\label{app:catalog}

\begin{figure}[htbp]
\centering
\includegraphics[width=\linewidth]{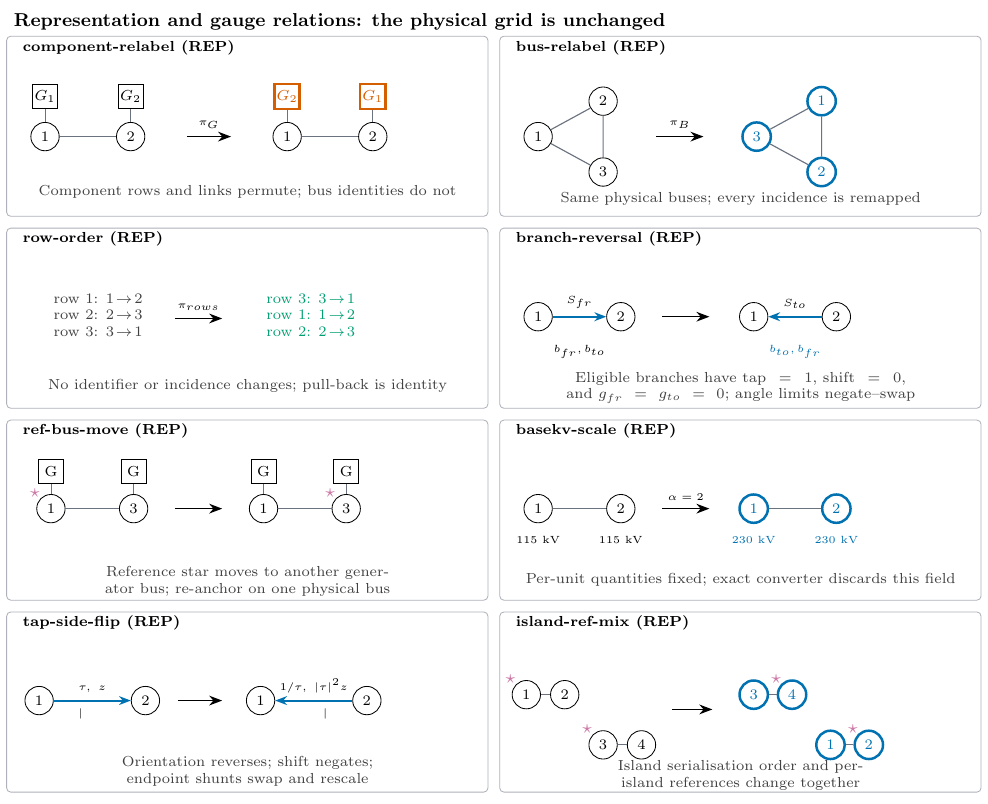}
\caption{Examples of \REP{} relations. They change labels, row order, branch orientation, or angle reference \crrev{without changing the underlying AC-OPF problem, and we compare predictions after undoing the declared map.} The pipeline-specific \texttt{basekv-\allowbreak scale} relation changes an exported model input that is absent from the standard per-unit AC-OPF target.}
\label{fig:relation-atlas-rep}
\end{figure}

\begin{figure}[htbp]
\centering
\includegraphics[width=\linewidth]{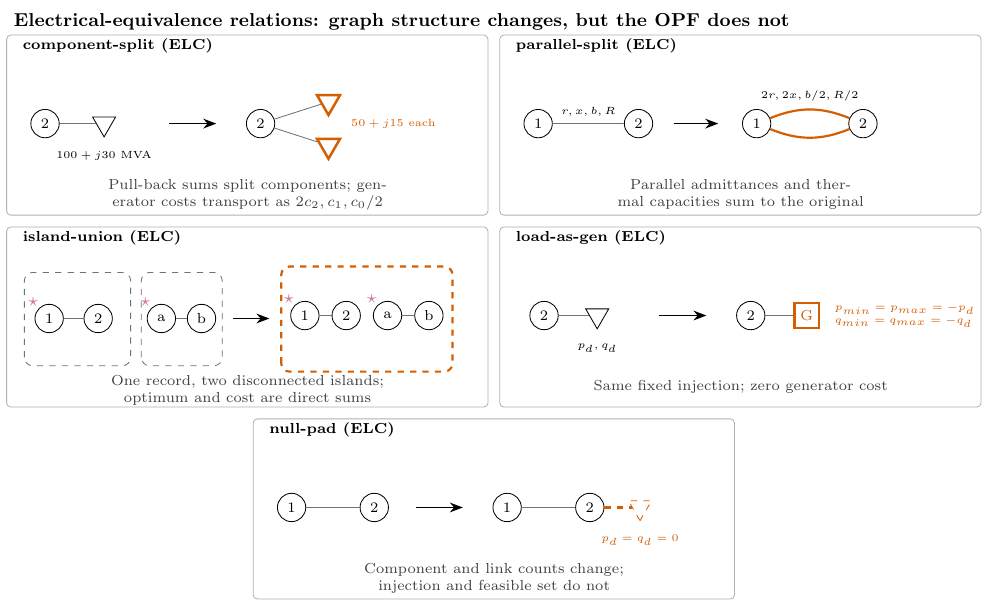}
\caption{Examples of \ELC{} relations. They change graph size or component-family bookkeeping while preserving the AC-OPF feasible set and objective under the stated validity conditions. Their output maps sum split quantities, separate joined islands, \crrev{reassign a fixed injection, or remove inert padding, depending on the relation.}}
\label{fig:relation-atlas-elc}
\end{figure}

\crrev{\Cref{tab:relations} defines each transformation, states its validity conditions, and gives the output map that returns predictions to the original physical quantities. \Cref{fig:relation-atlas-rep,fig:relation-atlas-elc} illustrate the two tiers: \REP{} relations change representation or angle reference, while \ELC{} relations change graph or component bookkeeping and require aggregation or an inverse map.}

{\scriptsize
\begin{longtable}{@{}p{1.7cm}p{0.7cm}p{4.2cm}p{2.6cm}p{3.0cm}@{}}
\caption{Catalog of the 13 metamorphic relations. For each relation, $T_r$ constructs an equivalent input and $P_r$ aligns or aggregates the transformed prediction with the original physical quantities; the validity condition states when that equivalence holds.}\label{tab:relations}\\
\toprule
Relation & Tier & {Transformation $T_r$} & Validity condition & {Output map $P_r$} \\
\midrule
\endfirsthead
\multicolumn{5}{c}{\tablename\ \thetable\ (continued)}\\
\toprule
Relation & Tier & {Transformation $T_r$} & Validity condition & {Output map $P_r$} \\
\midrule
\endhead
\midrule
\multicolumn{5}{r}{Continued on next page}\\
\endfoot
\bottomrule
\endlastfoot
component-relabel & \REP & Reorder the generator array (or load, or shunt) by a permutation $\sigma$, moving its bus-link rows with it; buses and branches untouched. & always (every $\sigma$) & apply $\sigma^{-1}$ to the per-component outputs \\
\addlinespace
bus-relabel & \REP & Reorder the bus array by a permutation $\pi$; rewrite every stored bus index through $\pi$ (branch from/to endpoints and all component-to-bus links). & always (every $\pi$) & apply $\pi^{-1}$ to the per-bus outputs \\
\addlinespace
row-order & \REP & Reorder the rows of one edge family (\texttt{ac\_line}, transformer, or a component-link table); every field value and endpoint unchanged (a pure serialisation reshuffle). & always & identity on physical quantities; re-sort through \texttt{canonical\_\allowbreak link\_\allowbreak order} before positional conversion \\
\addlinespace
branch-reversal & \REP & Reverse a branch: swap its from/to endpoints, swap the shunt halves $b_{fr}\!\leftrightarrow\!b_{to}$, and replace $(\text{angmin},\text{angmax})$ with $(-\text{angmax},-\text{angmin})$. & per row: $\tau{=}1$, shift ${=}0$, $g_{fr}{=}g_{to}{=}0$ (all \texttt{ac\_line} rows; only untapped, unshifted transformers) & reverse that branch's directed-flow channels, $(P_f,Q_f,P_t,Q_t)\!\to\!(P_t,Q_t,P_f,Q_f)$ \\
\addlinespace
ref-bus-move & \REP & Move the angle reference: set a chosen generator bus's type to reference and demote the former reference to PV; no other field changes. & the chosen bus is a generator bus in the same island (one reference per island) & identity on components; re-anchor both solutions' angles to a common physical bus \\
\addlinespace
basekv-scale & \REP & {{Multiply the model-visible \texttt{base\_kv} field at every bus by a common factor $\alpha>0$ after OPFData export.}} & {{pipeline-specific: the standard per-unit AC-OPF target is unchanged}} & {identity (solver quantities are unchanged)} \\
\addlinespace
tap-side-flip & \REP & {Swap a transformer's from/to endpoints. For complex tap $\tau{=}t e^{\mathrm{j}\phi}$, set $\tau'{=}\tau^{-1}$, $z'{=}t^2z$, $(b'_{fr},b'_{to}){=}(b_{to}/t^2,b_{fr}/t^2)$, and $(\theta'_{\min},\theta'_{\max}){=}(-\theta_{\max},-\theta_{\min})$.} & {$t>0$; $Y'{=}\Pi Y\Pi$ for the endpoint-swap matrix $\Pi$} & {reverse directed-flow channels, $(P_f,Q_f,P_t,Q_t)\!\to\!(P_t,Q_t,P_f,Q_f)$} \\
\addlinespace
island-ref-mix & \REP & Multi-island only: permute which island serialises first and choose each island's reference independently (bus relabeling on island order composed with per-island ref-bus-move). & multi-island instance; one valid generator reference per island & re-anchor each island's angles, then undo the island/bus permutation \\
\addlinespace
component-split & \ELC & {Split load $\ell$ into co-located loads $\ell_1,\ell_2$ satisfying $(P_{\ell_1}^d{+}P_{\ell_2}^d,Q_{\ell_1}^d{+}Q_{\ell_2}^d){=}(P_\ell^d,Q_\ell^d)$; or split generator $g$ into two with halved $(P_g^{\min},P_g^{\max},Q_g^{\min},Q_g^{\max})$ bounds and cost $(c_{g,2},c_{g,1},c_{g,0})\!\to\!(2c_{g,2},c_{g,1},c_{g,0}/2)$ each.} & {loads: always; generators: convex quadratic cost and symmetric halved bounds} & \emph{sum} the two parts' injections back (aggregation, not a permutation) \\
\addlinespace
parallel-split & \ELC & Replace one branch by two identical parallel branches: $r,x\!\to\!2r,2x$ (parallel admittance preserved), $b_{fr},b_{to}\!\to\!b_{fr}/2,b_{to}/2$, rate $\to$ rate$/2$; angle limits kept. & always (admittances recombine; flow splits evenly) & sum the two branches' flows \\
\addlinespace
island-union & \ELC & Concatenate two solved scenarios into one instance with two disconnected islands, renumbering the second scenario's buses and components after the first. & always (no coupling; optimum and cost are the direct sum) & split the joint prediction into the two island blocks \\
\addlinespace
load-as-gen & \ELC & {Recast a fixed load $\ell$ with demand $(P_\ell^d,Q_\ell^d)$ as a zero-cost generator $g'$ at the same bus with $P_{g'}^{\min}{=}P_{g'}^{\max}{=}{-}P_\ell^d$ and $Q_{g'}^{\min}{=}Q_{g'}^{\max}{=}{-}Q_\ell^d$, then remove $\ell$ from the load table.} & always (a clamped zero-cost generator imposes the identical injection) & {read $(P_{g'},Q_{g'})$ and map it back to the load (must equal $(-P_\ell^d,-Q_\ell^d)$)} \\
\addlinespace
null-pad & \ELC & {Add an inert element: a load $\ell'$ with $P_{\ell'}^d{=}Q_{\ell'}^d{=}0$, a shunt with $G^{\mathrm{sh}}{=}B^{\mathrm{sh}}{=}0$, or a zero-cost generator $g'$ with $P_{g'}^{\min}{=}P_{g'}^{\max}{=}Q_{g'}^{\min}{=}Q_{g'}^{\max}{=}0$; bus types are unchanged.} & always (the element injects nothing; only counts change) & drop the padded element from the prediction \\
\end{longtable}
}

\section{Technical foundation}
\label{app:technical}

\subsection{AC optimal power flow and non-uniqueness}
\label{app:acopf-nonuniqueness}

Let $\mathcal N$, $\mathcal E$, and $\mathcal G$ denote buses, branches, and generators. For bus $i$, the AC-OPF variables include voltage magnitude $V_i$, voltage angle $\theta_i$, and active and reactive generation $(P_g,Q_g)$; fixed demands are $(P_i^d,Q_i^d)$. With commitment fixed, a common economic-dispatch objective is
\begin{equation}
\min_{V,\theta,P,Q}\quad \sum_{g\in\mathcal G} c_{g,2}P_g^2+c_{g,1}P_g+c_{g,0}.
\label{eq:appendix-acopf-objective}
\end{equation}
Writing $Y=G+\mathrm{j}B$ and $\theta_{ij}=\theta_i-\theta_j$, active and reactive nodal balance require
\begin{crrevision}
\begin{align}
\sum_{g\in\mathcal G_i}P_g-P_i^d &= V_i\sum_{j\in\mathcal N}V_j\left(G_{ij}\cos\theta_{ij}+B_{ij}\sin\theta_{ij}\right),
\label{eq:appendix-acopf-active-balance}\\
\sum_{g\in\mathcal G_i}Q_g-Q_i^d &= V_i\sum_{j\in\mathcal N}V_j\left(G_{ij}\sin\theta_{ij}-B_{ij}\cos\theta_{ij}\right).
\label{eq:appendix-acopf-reactive-balance}
\end{align}
\end{crrevision}

Generator outputs and voltage magnitudes obey bounds; branch flows obey apparent-power and angle-difference limits. One reference angle per connected island removes the additive angle gauge. \crrev{Voltage products and trigonometric angle differences make the feasible set nonconvex. AC feasibility is strongly NP-hard in general, and multiple local optima can occur \citep{bienstockverma2019strong,bukhshetal2013local}. Production cost may also leave reactive dispatch and voltage magnitude weakly identified. Our test checks whether one deterministic surrogate is consistent across equivalent inputs and does not assume a unique AC-OPF optimum.}

\subsection{Fibers of the data-to-problem map}
\label{app:data-fibers}

\crrev{Serialized cases include choices that the physical optimization does not depend on, such as bus names, array order, stored branch direction, and angle reference. The problem itself depends on aggregate bus injections, series and shunt admittances, transformer ratios, incidence, variable bounds, and cost coefficients.} Let $s:\mathcal D\to\mathcal Z$ map serialized data to these sufficient quantities. A \emph{fiber} of $s$ contains the encodings with the same sufficient quantities. Valid relations stay within such a fiber, allowing for label permutations or changes of angle reference. After mapping outputs back, a representation-consistent surrogate satisfies $P_r f(T_r d)=f(d)$, or $\Delta_r(d)=0$ in \eqref{eq:empirical-discrepancy}.

\begin{table}[htbp]
\caption{Schema distinctions discarded by the data-to-problem map $s$, with the relations that test each resulting many-to-one collapse.}
\label{tab:fibers}
  {%
\centering
\small
{
\begin{tabular}{@{}p{0.57\linewidth}p{0.36\linewidth}@{}}
\toprule
Collapse in $s$ & Relations \\
\midrule
Incidence is label-free & \texttt{component-relabel}, \texttt{bus-relabel}, \texttt{row-order}, \texttt{island-ref-mix} \\
Only the aggregate branch model enters the equations & \texttt{branch-reversal}, \texttt{tap-side-flip} \\
Bus balance sees aggregate injection instead of per-device & \texttt{component-split}, \texttt{load-as-gen}, \texttt{null-pad} \\
Parallel branches collapse to one effective admittance & \texttt{parallel-split} \\
{Angle reference is a gauge convention} & {\texttt{ref-bus-move}, \texttt{island-ref-mix}} \\
{The post-export per-unit target does not use model-visible \texttt{base\_kv}} & {\texttt{basekv-scale}} \\
Disconnected islands cannot interact & \texttt{island-union} \\
\bottomrule
\end{tabular}
}
}
\end{table}

Each row of \cref{tab:fibers} identifies a serialization choice that leaves the optimization problem unchanged. The catalog tests one concrete transformation within each listed fiber; \cref{app:fiber-coverage} states the limits of this coverage.

\subsection{Fiber coverage and extensions}
\label{app:fiber-coverage}

\crrev{A sufficiently small relaxation of a strictly inactive bound could extend the catalog: under local solution stability, it preserves the same local solution, although it need not preserve the global OPF problem.} Unequal or $N$-way component splits and arbitrary shunt redistribution would extend the discrete tests to continuous families.

\crrev{Coverage remains incomplete even for the listed fibers. Most are continuous, so one discrete transformation tests only part of each. For example, \texttt{branch-reversal} swaps endpoints without testing arbitrary redistribution of $b_{fr}+b_{to}$, and \texttt{component-split} uses equal halves rather than an arbitrary $N$-way split. We have not shown that compositions of the 13 relations generate every fiber. The catalog is therefore representative but not exhaustive.}

\subsection{When augmentation theory applies}
\label{app:augmentation-theory}

\crrev{For squared loss, orbit averaging projects a predictor onto the equivariant function class when the transformations form a compact group, the target is equivariant, and the input distribution is invariant or orbit-balanced. Under these assumptions, the non-equivariant component contributes avoidable orbit-balanced risk \citep{elesedyzaidi2021provably,chenetal2020grouptheoretic,lyleetal2020invariance}.} Sampling transformations from the normalized Haar measure gives an unbiased stochastic estimate of this orbit-averaged objective.

Only part of the catalog satisfies these assumptions. \crrev{Finite relabelings and eligible involutions fit the compact-group setting. By contrast, \texttt{basekv-scale} is unbounded and pipeline-specific, reference changes are conditional gauge actions, and \ELC{} relations can change graph and output dimensions. The model-specific $0.5$ mixture over eligible relations therefore does not sample a single group's Haar measure. The theorem concerns orbit-balanced risk, not error on one canonical serialization, and nonconvex training need not reach the projection \citep{nordenforsetal2024optimization}.} We report canonical accuracy, transformed accuracy, metamorphic discrepancy, and feasibility separately.

\section{Audit design and model inventory}
\label{app:model-inventory}


\Cref{tab:screen-models} lists every architecture and mechanism in the built-in consistency screen described in \cref{sec:method}. \crrev{The reimplementations test five design choices: output partitions based on array index versus slack/PV/PQ role; raw Laplacian coordinates; summaries over index-selected landmarks versus all buses; learned per-index versus graph-fixed filters; and directed versus orientation-symmetric branch processing.} Where practical, matched pairs use the same surrounding network and random seed and change only the component of interest. They isolate that component's effect but do not reproduce a published model's training or accuracy.

We also include two MPNN rows that connect the architecture screen to the augmentation study. \crrev{They are randomly initialized instances of the study architectures, each with four message-passing layers, width 128, sum aggregation, orientation-invariant branch features, and masked solution inputs; the second row adds 16 raw Laplacian eigenvectors.}
The three plain MPNNs provide smaller family controls for mean versus sum aggregation and for branch orientation.

\crrev{A random-weight failure shows that the tested pipeline does not guarantee invariance to that relation; it does not establish trained accuracy, feasibility, or discrepancy magnitude. The DC3 rows test only the partition-and-completion choice with a stand-in. They do not run DC3's Newton power-flow completion or correction loop.}

{\scriptsize
\begin{longtable}{@{}p{3.2cm}p{2.0cm}p{7.0cm}@{}}
\caption{Inventory of the 17 model, reimplementation, and control rows in the built-in consistency screen. \crrev{``Scope'' names the released pipeline or design mechanism behind each row.}}\label{tab:screen-models}\\
\toprule
Row & Class & Scope \\
\midrule
\endfirsthead
\multicolumn{3}{c}{\tablename\ \thetable\ (continued)}\\
\toprule
Row & Class & Scope \\
\midrule
\endhead
\midrule
\multicolumn{3}{r}{Continued on next page}\\
\endfoot
\bottomrule
\endlastfoot
{GridSFM-Open v1.1} & {Open weight} & {Released 15,148,227-parameter GridSFM graph-transformer checkpoint and inference pipeline \citep{yangetal2026gridsfm,microsoftgridsfmopen2026modelcard,microsoftgridsfm2026software}.} \\
{LUMINA-2M v0.1.3} & {Open weight} & {Released 2,375,548-parameter, six-layer heterogeneous model and inference pipeline \citep{lietal2026lumina,argonnelumina2m2026modelcard}.} \\
{GridFM-GNS v0.8.1} & Public architecture & {Released heterogeneous preprocessing, model, and decoder at random initialization; we found no compatible public AC-OPF checkpoint as of August 25, 2026 \citep{gridfmgraphkit2026v081}.} \\
DC3 (index partition) & Design reimplementation & Random index masks for partial $V_m,V_a,P_g$ outputs followed by covariant mean completion \citep{dontietal2021dc3}. \\
DC3 (physical partition) & Matched control & Same completion stand-in with masks keyed by slack/PV/PQ roles rather than array position. \\
Graph transformer (LapPE-4) & Design reimplementation & Global bus self-attention with four raw, uncanonicalized Laplacian eigenvectors \citep{rampaseketal2022recipea}. \\
Landmark PE ($K{=}64$) & Mechanism control & Four shortest-path-distance moments to the first 64 bus indices, appended to an MPNN. \\
Landmark PE ($K{=}$all) & Matched control & Same moments over all buses, removing the index-selected subset. \\
Effective resistance ($K{=}64$) & Design reimplementation & Five unit-weight effective-resistance moments over the first 64 bus indices, appended to an MPNN \citep{microsoftgridsfm2026software,yangetal2026gridsfm}. \\
Effective resistance ($K{=}$all) & Matched control & Same moments over all buses \citep{arowolocremer2026generalization}. Both effective-resistance rows use unit edge weights rather than the susceptance weighting of the cited designs, isolating index selection from electrical weighting. \\
Index filter (fixed) & Matched control & Three bilinear graph-filter layers using degree-normalized adjacency directly. \\
Index filter (learned) & Design reimplementation & Same layers with learned per-grid $N\times N$ values masked to graph support \citep{liuetal2023topologyaware}. \\
MPNN (matched) & Matched control & Four-layer, width-128 heterogeneous MPNN with sum aggregation, symmetric branch features, and masked solution inputs. \\
MPNN + LapPE (matched) & Matched control & Same MPNN with 16 raw, uncanonicalized Laplacian eigenvectors. \\
{Plain MPNN (mean)} & {Family control} & {Two-layer heterogeneous MPNN with shared encoders and heads, mean aggregation, and symmetric branch encoding; this control is inspired by OPF-HGNN but does not reproduce its exact aggregation or branch encoding \citep{ghamizietal2024opfhgnn}.} \\
Plain MPNN (sum) & Family control & Same two-layer MPNN with sum aggregation. \\
Plain MPNN (orientation naive) & Family control & Same sum MPNN with raw endpoint fields and only the stored branch direction. \\
\end{longtable}
}

\subsection{Data, sampling, and solver admission}
\label{app:audit-controls}

\begin{crrevision}
Scenarios come from the N-1 topology-perturbed OPFData release; both released checkpoints were trained on this type of perturbation \citep{lovettetal2024opfdata,microsoftgridsfmopen2026modelcard,argonnelumina2m2026modelcard}. LUMINA metadata includes all five audited grids. GridSFM documents a training-size range of 500 to 4,661 buses, placing case14/118 outside that range and the larger grids inside it. These scenarios therefore do not provide a common unseen-topology benchmark for the two checkpoints.

The architecture screen uses case14/500, three source scenarios per grid, three draws per applicable case--relation pair, and three initializations where applicable. Fourteen \texttt{component-split} cells retain only case500 evidence because the case14 control was unavailable. Source scenarios are the independent evaluation units; transformed views of one source are repeated measurements.

\textbf{Five-grid comparison sampling.} The separate trained-versus-random audit uses three source scenarios from each of case14, case118, case500, case2000, and case4661, with three sampled transformations per source--relation pair. The same source scenarios and transformation draws are used for each backbone's released checkpoint and its three random initializations (seeds 0, 1, and 2). Each grid and model/seed run therefore contains $3\times13\times3=117$ executed source--relation--draw comparisons before aggregation. The trained-checkpoint runs stored under three seed labels reuse the same checkpoint and draws; these repetitions do not increase the independent sample count. The independent units remain the three source scenarios per grid.

\textbf{What the solver controls establish.} PowerModels.jl and IPOPT check the numerical implementation of the relations \citep{coffrinetal2018powermodels,wachterbiegler2005implementation}. The algebraic conditions still determine validity, and a converged solve need not be globally optimal. The archived expanded-relation configuration uses the polar AC formulation (\texttt{ACPPowerModel}), a flat start in each view, an IPOPT tolerance of $10^{-8}$, and a 300-second limit. The wrapper accepts its recorded converged statuses: \texttt{LOCALLY\_SOLVED}, \texttt{OPTIMAL}, and their \texttt{ALMOST} variants. These labels do not independently verify residual bounds or global optimality. The saved control summaries do not pin the linear solver.

\textbf{Expanded-relation admission.} The historical control implementation admits a grid--relation pair only if all its sampled solves converge, the worst relative objective difference is at most $10^{-9}$, and the worst aligned primal difference meets its relation-specific limit. Relative objective error is $|C'-C_{\mathrm{ref}}|/\max(|C_{\mathrm{ref}}|,10^{-12})$; disconnected-union controls use the sum of component reference costs. The primal limits are $0$ for \texttt{basekv-scale}, $10^{-9}$ for \texttt{ref-bus-move}/\texttt{load-as-gen}, $10^{-7}$ for \texttt{null-pad}, $10^{-6}$ for \texttt{component-split}/\texttt{parallel-split}, $2.2\times10^{-6}$ for \texttt{tap-side-flip}, and $10^{-5}$ for \texttt{island-union}/\texttt{island-ref-mix}.

The archived case14/500 table contains 18 grid--relation pairs with nine control records each. All 162 pairs of solves converge, and 17 grid--relation pairs pass admission. Case14 \texttt{component-split} is unavailable because its objective difference, $7.26\times10^{-9}$, exceeds $10^{-9}$, even though IPOPT converged. The model screen reuses these controls and does not run a fresh solve for every model draw. Although the configuration names $10^{-6}$ equality/inequality targets, the expanded control CSV does not store independently audited residuals, so we do not claim that those targets were separately verified for every admitted view.

\textbf{Original relabeling and orientation controls.} For the four original relations, we check transported labels and residual preservation over 300 records per grid--relation on case14/118/500 and 90 on case2000/4661. We also check bus relabeling against an independent converter. Transported labels match, and residual changes meet $10^{-12}$ in all 20 grid--relation summaries. A separate re-solve comparison uses $10^{-6}$ cost and primal tolerances. The case4661 row-order and branch-reversal controls converge but exceed the primal threshold. We retain these outcomes as sensitivity of the solver solution, rather than reclassifying them as surrogate failures. This distinction matters when solutions are non-unique or weakly identified.

\textbf{Verdict versus descriptive magnitude.} The screen takes the largest channel-mean discrepancy over all executed draws and compares it with $10^{-6}$; the separate maximum-component check uses $10^{-5}$. The five-grid comparison first averages draws within each source, then takes the median across sources and the maximum across grids, retaining a value for each of three random seeds. Its checkmarks mean that this aggregate is below threshold, even if some individual draws exceed it. Neither finite screen passes nor below-threshold aggregates prove universal invariance.
\end{crrevision}

\begin{crrevision}
\subsection{Threshold sensitivity and component tails}
\label{app:screen-sensitivity}
We reaggregate the 221 saved model--relation summaries from the 17-row screen, changing only the reporting threshold and keeping the executed samples and applicability mask fixed. No model or solver is rerun. Every configuration still fails at least six relations at $10^{-7}$, $10^{-6}$, and $10^{-5}$ in native channel units. At the primary $10^{-6}$ channel-mean threshold, all 64 passing cells also have saved maximum-component discrepancies at or below $10^{-5}$. The component check therefore reveals no additional failure among the screen passes. This result does not establish an all-draw pass for the separate five-grid aggregate comparison.
\begin{table}[htbp]
\crcolor
\caption{Sensitivity of the saved architecture-screen verdicts to the channel-mean reporting threshold. Each row uses the same 221 cells and partial-coverage mask.}
\label{tab:screen-threshold-sensitivity}
\centering\small
\begin{tabular}{@{}lrrr@{}}
\toprule
Threshold & Failed cells / 221 & Failed relations per model & Models with a failure \\
\midrule
$10^{-7}$ & 160 & 6--13 & 17/17 \\
$10^{-6}$ & 157 & 6--13 & 17/17 \\
$10^{-5}$ & 154 & 6--12 & 17/17 \\
$10^{-4}$ & 144 & 4--12 & 17/17 \\
$10^{-3}$ & 134 & 4--12 & 17/17 \\
\bottomrule
\end{tabular}
\end{table}

\crrev{Excluding \texttt{basekv-scale} leaves 204 model--relation cells. At the primary $10^{-6}$ threshold, 140 of these cells fail, and every configuration fails between five and twelve of the twelve remaining relations. At $10^{-4}$ and $10^{-3}$, respectively, 127 and 118 cells fail; every configuration still fails between three and eleven of the twelve remaining relations.}

The tolerance is a prespecified reporting threshold, not an empirically calibrated noise floor for every pipeline. The saved results lack a comprehensive set of repeated-forward, round-trip, CPU/GPU, and float32/float64 comparisons. Near-threshold cells therefore cannot identify a particular numerical cause. The table shows that the broad failure-count conclusion survives changes to the reporting threshold; it does not establish robustness to changes in arithmetic precision.
\end{crrevision}

\section{Detailed audit evidence}
\label{app:weight-study}

\subsection{Trained versus random-weight consistency}
\label{app:trained-random}

\crrev{\Cref{tab:trained-random-summary} uses the worst-grid summary defined in \cref{app:audit-controls}. Each trained entry uses the released checkpoint; the random entry gives the range across three seed-specific aggregates. Checkmarks classify the aggregate, not every draw.}

\begin{table}[htbp]
\caption{Trained and random-weight worst-grid discrepancies by relation. Entries are unnormalized worst-channel mean absolute discrepancies; random entries span three weight seeds. \crrev{A checkmark marks an aggregate value at or below the prespecified $10^{-6}$ channel tolerance; it does not mean that every draw passes.} Because values inherit the worst channel's unit (p.u. for $V_m,P_g,Q_g$ or radians for $V_a$), they are comparable with the tolerance but not across channels.}
\label{tab:trained-random-summary}
  {%
\centering
\scriptsize
\setlength{\tabcolsep}{3.5pt}
\renewcommand{\arraystretch}{1.04}
{
\begin{tabular}{@{}lrrrr@{}}
\toprule
& \multicolumn{2}{c}{GridSFM-Open v1.1} & \multicolumn{2}{c}{LUMINA-2M v0.1.3} \\
\cmidrule(lr){2-3}\cmidrule(l){4-5}
Relation & Trained & Random seed range & Trained & Random seed range \\
\midrule
\texttt{component-relabel} & $3.80\times10^{-7}\,\checkmark$ & $[3.00\times10^{-8},\,3.53\times10^{-8}]\,\checkmark$ & $2.18\times10^{-8}\,\checkmark$ & $[4.07\times10^{-9},\,1.03\times10^{-8}]\,\checkmark$ \\
\texttt{bus-relabel} & $0.279$ & $[0.00957,\,0.0143]$ & $5.01\times10^{-9}\,\checkmark$ & $[5.60\times10^{-10},\,3.37\times10^{-9}]\,\checkmark$ \\
\texttt{row-order} & $0.780$ & $[0.574,\,0.578]$ & $4.45\times10^{-8}\,\checkmark$ & $[5.55\times10^{-9},\,1.17\times10^{-8}]\,\checkmark$ \\
\texttt{branch-reversal} & $0.326$ & $[0.0133,\,0.0365]$ & $0.299$ & $[0.00528,\,0.00903]$ \\
\texttt{ref-bus-move} & $0.438$ & $[0.00755,\,0.0113]$ & $0.182$ & $[2.40\times10^{-4},\,0.00789]$ \\
\texttt{component-split} & $0.00940$ & $[3.45\times10^{-4},\,0.00232]$ & $0.00325$ & $[2.45\times10^{-5},\,7.04\times10^{-5}]$ \\
\texttt{parallel-split} & $0.0253$ & $[0.00171,\,0.00206]$ & $1.15\times10^{-4}$ & $[2.86\times10^{-5},\,6.86\times10^{-5}]$ \\
\texttt{island-union} & $0.680$ & $[0.0515,\,0.160]$ & $3.30\times10^{-8}\,\checkmark$ & $[7.51\times10^{-9},\,2.23\times10^{-8}]\,\checkmark$ \\
\texttt{basekv-scale} & $0.397$ & $[0.0168,\,0.0232]$ & $0.513$ & $[0.0547,\,0.0782]$ \\
\texttt{tap-side-flip} & $0.104$ & $[0.00569,\,0.0141]$ & $0.0333$ & $[1.47\times10^{-4},\,2.02\times10^{-4}]$ \\
\texttt{load-as-gen} & $0.0322$ & $[0.00782,\,0.0175]$ & $0.00157$ & $[3.12\times10^{-5},\,2.06\times10^{-4}]$ \\
\texttt{null-pad} & $0.104$ & $[0.00248,\,0.00767]$ & $0.0231$ & $[8.96\times10^{-4},\,0.00226]$ \\
\texttt{island-ref-mix} & $0.675$ & $[0.244,\,0.268]$ & $0.124$ & $[2.40\times10^{-4},\,0.00789]$ \\
\bottomrule
\end{tabular}
}
}
\end{table}

\begin{figure}[htbp]
\centering
\includegraphics[width=0.72\linewidth]{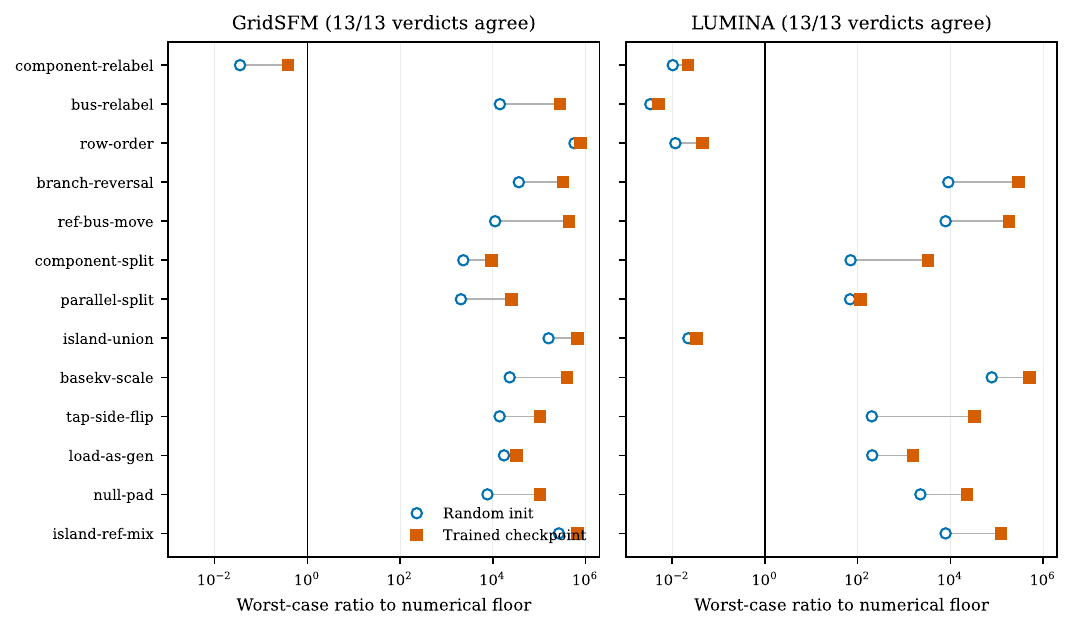}
\caption{\crrev{Trained and random-weight models agree on all 13 threshold classifications of the displayed aggregate statistic.} \crrev{For each relation, the horizontal coordinate is the worst-channel discrepancy, maximized over five grids, divided by the $10^{-6}$ tolerance; random-weight markers also take the worst of three seeds. \crrev{The vertical line at one marks the aggregate reporting threshold and does not serve as an all-draw pass criterion.}} Connecting segments show the change in magnitude after training.}
\label{fig:weight-study-verdict}
\end{figure}

\begin{figure}[htbp]
\centering
\includegraphics[width=0.92\linewidth]{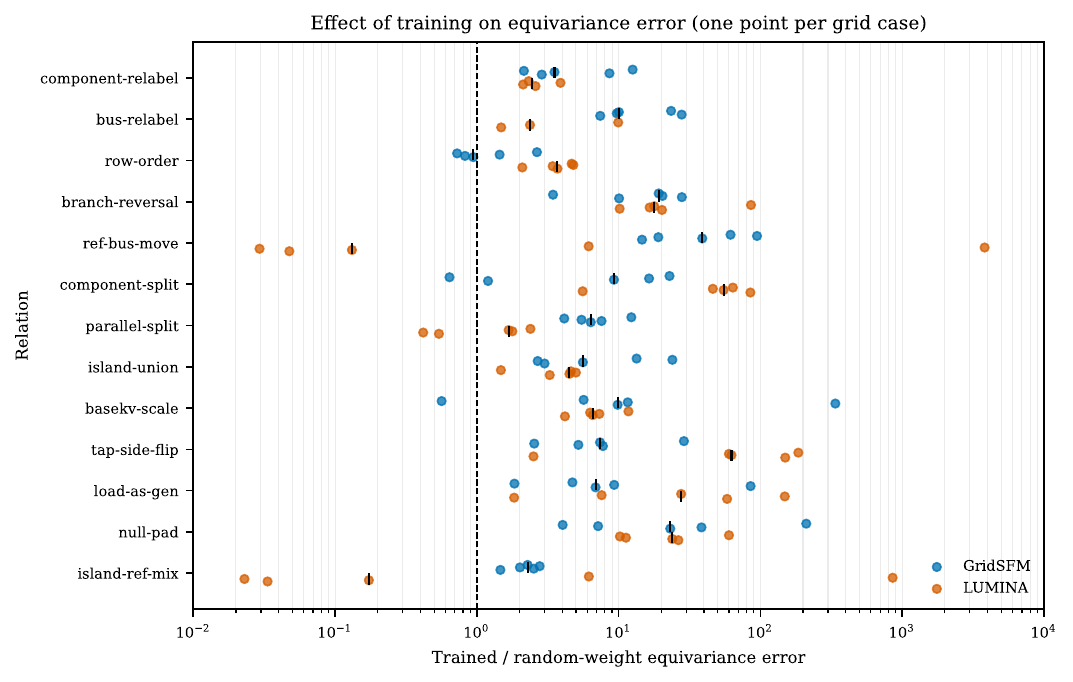}
\caption{Per-grid effect of training on metamorphic discrepancy. Each point is the trained discrepancy divided by its random-weight control for one grid and relation; values above one indicate a larger mismatch after training. Short black bars mark relation-level medians. \crrev{Ratios for \texttt{component-relabel} compare values that are already below the $10^{-6}$ tolerance, so they carry little information.}}
\label{fig:weight-study-ratio}
\end{figure}

\begin{figure}[htbp]
\centering
\includegraphics[width=0.92\linewidth]{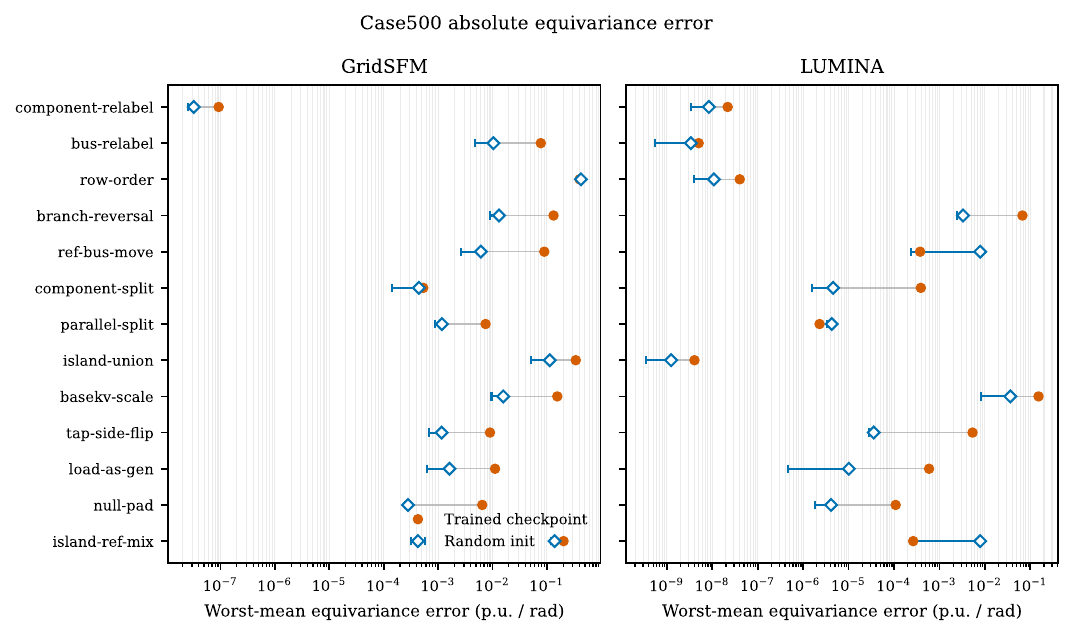}
\caption{Absolute case500 discrepancies underlying the ratio comparison. Circles show released checkpoints; diamonds and intervals show the central value and range across three random seeds. The plot distinguishes a large relative change near zero from a discrepancy far above $10^{-6}$. \crrev{Each point uses its worst channel, so values should be read against the tolerance and not compared across channel units.}}
\label{fig:weight-study-absolute}
\end{figure}

\crrev{\Cref{tab:trained-random-summary,fig:weight-study-verdict,fig:weight-study-ratio,fig:weight-study-absolute} show the same aggregate threshold classifications for trained and random weights. GridSFM is below threshold only on \texttt{component-relabel}; LUMINA is also below it on \texttt{bus-relabel}, \texttt{row-order}, and \texttt{island-union}. Excluding \texttt{component-relabel}, whose trained and random discrepancies are below tolerance, training increases discrepancy in 55 of 60 finite GridSFM case--relation cells and 50 of 58 finite LUMINA cells. The median ratios are $7.50$ and $6.23$, respectively. Ratios must be read alongside absolute discrepancies: a near-zero random denominator can produce a large ratio, and a ratio below one can still leave a discrepancy above tolerance. Random networks may also be nearly constant. These comparisons measure representation sensitivity, not predictive quality; all trained and random arms fail the AC-feasibility criterion.}

\subsection{Auxiliary dispatch-cost sensitivity}
\label{app:dispatch-cost}

\begin{crrevision}
Let $\widehat P_g(d)=[f(d)]_{P_g}$ and $\widehat P_{g,r}(d)=[P_r f(T_r d)]_{P_g}$ be the canonical and aligned transformed active-power predictions. With the cost $C$ in \cref{eq:appendix-acopf-objective}, define
\begingroup
\setlength{\abovedisplayskip}{4pt}
\setlength{\belowdisplayskip}{4pt}
\begin{equation}
S_r(d)=\frac{\left|C\!\left(\widehat P_{g,r}(d)\right)-C\!\left(\widehat P_g(d)\right)\right|}{\max\!\left(\left|C\!\left(P_g^\star(d)\right)\right|,\varepsilon\right)},
\label{eq:dispatch-cost-sensitivity}
\end{equation}
\endgroup
where $P_g^\star(d)$ is the solver dispatch and $\varepsilon$ is machine epsilon. \Cref{tab:gridsfm-cost-movement} reports the median numerator in cost units and the median $S_r(d)$. For example, $S_r(d)=0.513$ means that an encoding change alters the implied cost by $51.3\%$ of the \crrev{solver's reference cost}. The predictions are AC infeasible, so $S_r(d)$ measures representation sensitivity rather than a realized dispatch loss or optimality gap. Random-weight values likewise do not measure economic quality.
\end{crrevision}
\begin{table}[htbp]
\caption{Median absolute difference between the costs implied by canonical and aligned transformed GridSFM predictions. Absolute values are in cost units; relative values divide that difference by the \crrev{solver's reference dispatch cost}. \crrev{Because the predictions are infeasible, these values measure sensitivity and are not optimality gaps.}}
\label{tab:gridsfm-cost-movement}
  {%
\centering
\small
\resizebox{\linewidth}{!}{%
\begin{tabular}{@{}lrrrr@{}}
\toprule
\multirow{2}{*}{Relation} & \multicolumn{2}{c}{Trained weights} & \multicolumn{2}{c}{Random weights} \\
\cmidrule(lr){2-3}\cmidrule(l){4-5}
& Absolute & Relative & Absolute & Relative \\
\midrule
\texttt{component-relabel} & $7.8\times10^{-3}$ & $1.57\times10^{-8}$ & $2.15\times10^{-3}$ & $8.18\times10^{-9}$ \\
\texttt{bus-relabel} & $1.25\times10^{4}$ & $0.022$ & $1.65\times10^{3}$ & $3.94\times10^{-3}$ \\
\texttt{row-order} & $6.99\times10^{4}$ & $0.234$ & $1.34\times10^{3}$ & $4.86\times10^{-3}$ \\
\texttt{branch-reversal} & $1.98\times10^{4}$ & $0.0404$ & $1.15\times10^{3}$ & $4.23\times10^{-3}$ \\
\texttt{ref-bus-move} & $2.60\times10^{4}$ & $0.0737$ & $281$ & $1.28\times10^{-3}$ \\
\texttt{component-split} & $57.5$ & $1.40\times10^{-4}$ & $2.04$ & $4.92\times10^{-5}$ \\
\texttt{parallel-split} & $180$ & $3.76\times10^{-4}$ & $45.2$ & $3.65\times10^{-4}$ \\
\texttt{island-union} & $4.63\times10^{4}$ & $0.502$ & $1.98\times10^{3}$ & $5.27\times10^{-3}$ \\
\texttt{basekv-scale} & $3.21\times10^{4}$ & $0.0350$ & $1.03\times10^{3}$ & $1.81\times10^{-3}$ \\
\texttt{tap-side-flip} & $366$ & $7.69\times10^{-4}$ & $105$ & $7.16\times10^{-4}$ \\
\texttt{load-as-gen} & $2.03\times10^{3}$ & $2.46\times10^{-3}$ & $66.8$ & $3.32\times10^{-4}$ \\
\texttt{null-pad} & $401$ & $1.08\times10^{-3}$ & $9.43$ & $9.71\times10^{-5}$ \\
\texttt{island-ref-mix} & $9.40\times10^{4}$ & $0.513$ & $4.22\times10^{3}$ & $0.0120$ \\
\bottomrule
\end{tabular}%
}
}
\end{table}
\clearpage

\section{Augmentation protocol and extended results}
\label{app:augmentation}

\subsection{Training and evaluation protocol}
\label{app:augmentation-protocol}

\begin{crrevision}
For each of four architectures, we train a baseline on canonical inputs and a second model with transformed inputs. The architectures are a four-layer, width-128 heterogeneous MPNN; that MPNN with 16 raw Laplacian eigenvectors (LapPE+MPNN); that MPNN with a 16-step random-walk structural encoding (RWSE+MPNN); and LUMINA. RWSE uses per-node return probabilities, avoiding eigenvector sign and basis ambiguities \citep{dwivedietal2023benchmarking,dwivedietal2022graph,rampaseketal2022recipea}. Each architecture uses three seeds (42, 31415, 271828), 2,000 case500 training instances, 89 epochs, 22,250 optimizer steps, batch size 8, and weight decay $10^{-4}$. Learning rates are $10^{-3}$ for the MPNN variants and $3\times10^{-4}$ for LUMINA. Each instance in the augmented arm is transformed by one eligible relation with probability $0.5$.

All arms minimize the same equal-channel MSE plus AC constraint penalties,
\begin{equation}
\mathcal{L}=\frac{1}{|\mathcal C|}\sum_{c\in\mathcal C}\operatorname{MSE}_c
+\lambda_{\mathrm{eq}}\log\!\Big(1+\frac1B\sum_b(r_{P,b}^2+r_{Q,b}^2)\Big)
+\lambda_{\mathrm{ineq}}\frac1K\sum_k\max(0,g_k)^2.
\label{eq:penalty-objective}
\end{equation}
Here $r_{P,b},r_{Q,b}$ are the real and reactive parts of $S^{\mathrm{inj}}-V\odot\overline{Y_{\mathrm{bus}}V}$; $g_k$ includes variable bounds, branch angle differences, and thermal limits at both ends. The frozen configuration sets $\lambda_{\mathrm{eq}}=3.606662\times10^{-4}$ and $\lambda_{\mathrm{ineq}}=5.929293\times10^{-3}$ for all 24 arms. These weights were calibrated offline using the median final validation-MSE-to-penalty ratios from nine earlier baseline arms on a 500-scenario validation split. \crrev{The recovered record reports all nine arms usable and reproduces both weights after rounding.} No one-epoch tuning procedure was involved. We select checkpoints using the validation objective. Within this experiment, the objective is fixed and augmentation changes; a comparison with the earlier pure-MSE experiments would change both the loss and the relation sets.

\textbf{Relation selection.} We fixed the training relation sets using discrepancies measured in earlier trained, unaugmented baseline arms and exclusions specific to each architecture. These sets did not come from the random-initialization screen. The seven MPNN and RWSE relations are \texttt{ref-bus-move}, \texttt{component-split}, \texttt{parallel-split}, \texttt{tap-side-flip}, \texttt{load-as-gen}, \texttt{null-pad}, and \texttt{island-ref-mix}. LapPE+MPNN adds \texttt{bus-relabel} and \texttt{island-union}. LUMINA uses the common seven plus \texttt{basekv-scale} and \texttt{branch-reversal}. The configuration records baseline discrepancies below $10^{-6}$ for \texttt{basekv-scale} in MPNN/LapPE and \texttt{island-union} in MPNN. Case500 features saturate, so these small discrepancies do not imply invariance for all inputs. Dashes in \cref{tab:augmentation-summary,tab:aug-appendix} mark relations excluded from training; evaluation still covers all thirteen relations.

\textbf{Evaluation cohorts.} The main evaluation uses 50 held-out case500 source scenarios per seed. Each has 53 views: one canonical input and four draws for each of thirteen relations. The 2,650 views are repeated measurements of 50 independent sources. A separate batched evaluation uses 200 canonical scenarios from each of case14, case118, case500, case2000, and case4661, without further training. None of its case500 scenario IDs occurs in the main cohort. \Cref{app:cohort-provenance} explains how the different cohorts and evaluation paths limit the comparison.

Accuracy averages the native-unit MAEs of $V_m,V_a,P_g,Q_g$ with equal weights. This convention combines per unit and radians; \cref{tab:augmentation-summary} reports the per-channel values needed to interpret it. For the main tables, improvement is the mean of three paired seed-level percentages, $100(e_{\mathrm{base},s}-e_{\mathrm{aug},s})/e_{\mathrm{base},s}$, so positive values mean lower error. \Cref{tab:aug-ood-clean} uses the opposite sign: negative values mean lower error. The spreads in \cref{app:augmentation-accuracy} are population standard deviations across the three percentage changes, not confidence intervals.

To summarize metamorphic consistency, we average sampled draws within each source, take the median across sources, then the median across the architecture's trained relations, and finally the mean over seeds. Per-relation entries instead use the seed with the largest source median. Models train on different relation sets, so these aggregate medians support baseline-versus-augmentation comparisons within a model but cannot rank architectures. Older evaluation artifacts used a $10^{-3}$ reporting scale. We recompute discrepancies from native-unit errors using the paper's $10^{-6}$ scale.

We assess AC feasibility separately. For each prediction, we first take the largest violation within each constraint family. Within a seed, we then take the worst of each source's 53 views and compute the median, linearly interpolated P95, and maximum over the 50 sources. Tables report the mean and sample standard deviation of the resulting three seed-level statistics. Equality violation is the larger of the maximum nodal active and reactive balance violations, in p.u. Pooled inequality violation is the maximum over constraint families after dividing each family's violation by its $10^{-6}$ native-unit tolerance; it is dimensionless. A source is feasible only if every view meets the equality and all inequality thresholds. No evaluated prediction meets these thresholds.
\end{crrevision}

\begin{table}[htbp]
\crcolor
\caption{{Full augmentation results under the AC constraint-penalty objective. Positive improvement (Imp.) denotes the mean seed-paired reduction under augmentation. Metamorphic values report the dimensionless normalized discrepancy $\Delta_r(d)$ defined in \eqref{eq:empirical-discrepancy}; $0$ denotes exact invariance and dashes mark excluded relations. Constraint rows use the scenario-clustered seed-level P95 defined in the text and report mean $\pm$ sample standard deviation across three seeds. Equality is in per unit. The pooled inequality metric is dimensionless and displayed in millions of family-specific $10^{-6}$ tolerances. Absolute family-level results appear in \cref{tab:penalty-violations-scaled}.}}
\label{tab:augmentation-summary}
\centering
\setlength{\tabcolsep}{2.0pt}
\renewcommand{\arraystretch}{1.00}
\resizebox{1.0\linewidth}{!}{%
{
\begin{tabular}{@{}lll*{4}{rrr}@{}}
\toprule
\multirow{2}{*}{Outcome} & \multirow{2}{*}{Presentation} & \multirow{2}{*}{Metric} & \multicolumn{3}{c}{MPNN} & \multicolumn{3}{c}{LapPE+MPNN} & \multicolumn{3}{c}{RWSE+MPNN} & \multicolumn{3}{c}{LUMINA} \\
\cmidrule(lr){4-6}\cmidrule(lr){7-9}\cmidrule(lr){10-12}\cmidrule(l){13-15}
 & & & Base & Aug. & Imp. (\%) & Base & Aug. & Imp. (\%) & Base & Aug. & Imp. (\%) & Base & Aug. & Imp. (\%) \\
\midrule
\multirow{10}{*}{Mean absolute error (MAE)} & \multirow{5}{*}{Canonical} & Mean & $0.01295$ & $0.01345$ & $-3.9$ & $0.09983$ & $0.01309$ & $86.8$ & $0.01255$ & $0.01250$ & $0.4$ & $0.01834$ & $0.01889$ & $-3.0$ \\
 & & $V_m$ & $0.00425$ & $0.00471$ & $-12.7$ & $0.01453$ & $0.00447$ & $69.1$ & $0.00407$ & $0.00424$ & $-4.2$ & $0.00674$ & $0.00758$ & $-12.4$ \\
 & & $V_a$ & $0.01289$ & $0.01290$ & $-0.1$ & $0.12493$ & $0.01320$ & $89.2$ & $0.01274$ & $0.01263$ & $0.9$ & $0.03500$ & $0.03606$ & $-3.0$ \\
 & & $P_g$ & $0.01198$ & $0.01268$ & $-5.9$ & $0.15529$ & $0.01176$ & $92.3$ & $0.01103$ & $0.01088$ & $1.4$ & $0.00952$ & $0.00961$ & $-0.9$ \\
 & & $Q_g$ & $0.02270$ & $0.02352$ & $-3.6$ & $0.10459$ & $0.02292$ & $77.9$ & $0.02236$ & $0.02226$ & $0.4$ & $0.02211$ & $0.02233$ & $-1.0$ \\
\cmidrule(lr){2-15}
 & \multirow{5}{*}{Transformed} & Mean & $0.01312$ & $0.01353$ & $-3.2$ & $0.10211$ & $0.01318$ & $87.0$ & $0.01271$ & $0.01257$ & $1.1$ & $0.03594$ & $0.02089$ & $41.9$ \\
 & & $V_m$ & $0.00427$ & $0.00472$ & $-12.5$ & $0.01468$ & $0.00447$ & $69.5$ & $0.00409$ & $0.00425$ & $-4.0$ & $0.00845$ & $0.00826$ & $2.2$ \\
 & & $V_a$ & $0.01308$ & $0.01300$ & $0.6$ & $0.13365$ & $0.01339$ & $89.9$ & $0.01293$ & $0.01271$ & $1.6$ & $0.06418$ & $0.04219$ & $34.3$ \\
 & & $P_g$ & $0.01227$ & $0.01279$ & $-4.2$ & $0.15438$ & $0.01187$ & $92.2$ & $0.01133$ & $0.01097$ & $3.1$ & $0.03263$ & $0.00977$ & $70.0$ \\
 & & $Q_g$ & $0.02285$ & $0.02361$ & $-3.4$ & $0.10573$ & $0.02299$ & $78.1$ & $0.02251$ & $0.02234$ & $0.7$ & $0.03850$ & $0.02333$ & $39.3$ \\
\midrule
\multirow{4}{*}{Out-of-distribution (OOD) MAE} & case14 & Mean & $0.29138$ & $0.27818$ & $3.8$ & $0.34591$ & $0.26954$ & $21.8$ & $0.29786$ & $0.28374$ & $4.3$ & $0.15930$ & $0.14110$ & $11.0$ \\
 & case118 & Mean & $0.47254$ & $0.53065$ & $-11.3$ & $0.48121$ & $0.46921$ & $2.8$ & $0.44662$ & $0.51278$ & $-13.5$ & $0.29558$ & $0.33423$ & $-14.2$ \\
 & case2000 & Mean & $0.25068$ & $0.24089$ & $3.9$ & $0.21860$ & $0.23550$ & $-7.7$ & $0.25316$ & $0.24444$ & $3.5$ & $0.22905$ & $0.21569$ & $5.7$ \\
 & case4661 & Mean & $0.49780$ & $0.49983$ & $-0.2$ & $0.41962$ & $0.47545$ & $-13.4$ & $0.47101$ & $0.49675$ & $-6.0$ & $0.48663$ & $0.44790$ & $8.0$ \\
\midrule
\multirow{12}{*}{Metamorphic} & Across trained relations & Median $\Delta_r$ & $852$ & $595$ & $30.1$ & $1.00\times10^{3}$ & $805$ & $16.2$ & $883$ & $563$ & $36.1$ & $4.63\times10^{3}$ & $1.26\times10^{3}$ & $70.8$ \\
\cmidrule(lr){2-15}
 & \multirow{11}{*}{Per relation} & \texttt{bus-relabel} & $0$ & $0$ & \textemdash & $1.95\times10^{5}$ & $5.32\times10^{3}$ & $97.3$ & $0$ & $0$ & \textemdash & $0$ & $0$ & \textemdash \\
 & & \texttt{branch-reversal} & \textemdash & \textemdash & \textemdash & \textemdash & \textemdash & \textemdash & \textemdash & \textemdash & \textemdash & $2.16\times10^{5}$ & $5.47\times10^{4}$ & $74.7$ \\
 & & \texttt{ref-bus-move} & $1.47\times10^{3}$ & $923$ & $37.1$ & $1.50\times10^{3}$ & $949$ & $36.9$ & $988$ & $1.02\times10^{3}$ & $-3.3$ & $8.02\times10^{4}$ & $6.29\times10^{4}$ & $21.5$ \\
 & & \texttt{component-split} & $898$ & $674$ & $24.9$ & $577$ & $668$ & $-15.7$ & $1.00\times10^{3}$ & $703$ & $29.9$ & $190$ & $139$ & $26.5$ \\
 & & \texttt{parallel-split} & $1.42\times10^{3}$ & $975$ & $31.5$ & $1.05\times10^{5}$ & $3.89\times10^{3}$ & $96.3$ & $1.25\times10^{3}$ & $999$ & $20.1$ & $23.8$ & $17.0$ & $28.6$ \\
 & & \texttt{island-union} & $0$ & $0$ & \textemdash & $1.62\times10^{5}$ & $5.81\times10^{3}$ & $96.4$ & $0$ & $0$ & \textemdash & \textemdash & \textemdash & \textemdash \\
 & & \texttt{basekv-scale} & \textemdash & \textemdash & \textemdash & \textemdash & \textemdash & \textemdash & \textemdash & \textemdash & \textemdash & $2.07\times10^{5}$ & $1.57\times10^{4}$ & $92.5$ \\
 & & \texttt{tap-side-flip} & $167$ & $90.9$ & $45.4$ & $61.9$ & $31.0$ & $49.9$ & $41.3$ & $50.0$ & $-21.2$ & $6.33\times10^{3}$ & $1.56\times10^{3}$ & $75.3$ \\
 & & \texttt{load-as-gen} & $1.01\times10^{3}$ & $644$ & $36.1$ & $658$ & $747$ & $-13.4$ & $1.19\times10^{3}$ & $594$ & $50.0$ & $1.98\times10^{3}$ & $478$ & $75.9$ \\
 & & \texttt{null-pad} & $1.32\times10^{3}$ & $736$ & $44.4$ & $803$ & $1.04\times10^{3}$ & $-29.4$ & $1.20\times10^{3}$ & $1.02\times10^{3}$ & $15.6$ & $797$ & $316$ & $60.4$ \\
 & & \texttt{island-ref-mix} & $1.10\times10^{3}$ & $695$ & $37.0$ & $1.59\times10^{5}$ & $5.74\times10^{3}$ & $96.4$ & $744$ & $766$ & $-2.9$ & $6.01\times10^{4}$ & $4.70\times10^{4}$ & $21.7$ \\
\midrule
\multirow{2}{*}{Constraint violations} & \multirow{2}{*}{All views} & Equality P95 (p.u.) & $58.1\!\pm\!18.7$ & $42.2\!\pm\!2.3$ & $22.1$ & $113.9\!\pm\!42.1$ & $39.5\!\pm\!5.2$ & $61.7$ & $51.8\!\pm\!10.9$ & $33.5\!\pm\!2.4$ & $33.4$ & $871\!\pm\!342$ & $164\!\pm\!11$ & $78.7$ \\
 & & Ineq. P95 ($10^6\!\times$ tol.) & $19.3\!\pm\!4.2$ & $11.9\!\pm\!1.6$ & $35.2$ & $42.1\!\pm\!4.6$ & $12.7\!\pm\!3.6$ & $70.1$ & $18.8\!\pm\!2.8$ & $10.2\!\pm\!0.8$ & $45.4$ & $276\!\pm\!91$ & $20.1\!\pm\!2.1$ & $92.3$ \\
\bottomrule
\end{tabular}
}}
\end{table}

\begin{table}[htbp]
\crcolor
\caption{{Expanded augmentation results under the AC constraint-penalty objective. Panel A compares three-seed mean accuracy, median $\Delta_r$ over each model's trained relations, and scenario-clustered constraint P95; violation entries are mean $\pm$ sample standard deviation across seeds. Equality is in per unit, while pooled inequality is dimensionless and displayed in millions of family-specific $10^{-6}$ tolerances. Feasible is the percentage of 50 source scenarios per seed whose 53 views all satisfy both $10^{-6}$ tolerances. Panel B gives the percentage reduction in $\Delta_r$ by relation; positive values denote improvement under augmentation, and dashes mark excluded relations.}}
\label{tab:aug-appendix}
  {%
\centering
\small
{\textbf{A. Aggregate outcomes}}\par\smallskip
\setlength{\tabcolsep}{4.5pt}
\renewcommand{\arraystretch}{1.08}
{
\begin{tabular}{@{}llrrrrrr@{}}
\toprule
Model & Arm & \shortstack{Canonical\\MAE} & \shortstack{Transformed\\MAE} & \shortstack{Median\\$\Delta_r$} & \shortstack{Equality P95\\(p.u.)} & \shortstack{Inequality P95\\($10^6\!\times$ tol.)} & \shortstack{Feasible\\(\%)} \\
\midrule
\multirow{2}{*}{MPNN} & Base & $0.01295$ & $0.01312$ & $852$ & $58.1\!\pm\!18.7$ & $19.3\!\pm\!4.2$ & $0$ \\
 & Aug. & $0.01345$ & $0.01353$ & $595$ & $42.2\!\pm\!2.3$ & $11.9\!\pm\!1.6$ & $0$ \\
\addlinespace
\multirow{2}{*}{LapPE+MPNN} & Base & $0.09983$ & $0.10211$ & $1.00\times10^{3}$ & $113.9\!\pm\!42.1$ & $42.1\!\pm\!4.6$ & $0$ \\
 & Aug. & $0.01309$ & $0.01318$ & $805$ & $39.5\!\pm\!5.2$ & $12.7\!\pm\!3.6$ & $0$ \\
\addlinespace
\multirow{2}{*}{RWSE+MPNN} & Base & $0.01255$ & $0.01271$ & $883$ & $51.8\!\pm\!10.9$ & $18.8\!\pm\!2.8$ & $0$ \\
 & Aug. & $0.01250$ & $0.01257$ & $563$ & $33.5\!\pm\!2.4$ & $10.2\!\pm\!0.8$ & $0$ \\
\addlinespace
\multirow{2}{*}{LUMINA} & Base & $0.01834$ & $0.03594$ & $4.63\times10^{3}$ & $871\!\pm\!342$ & $276\!\pm\!91$ & $0$ \\
 & Aug. & $0.01889$ & $0.02089$ & $1.26\times10^{3}$ & $164\!\pm\!11$ & $20.1\!\pm\!2.1$ & $0$ \\
\bottomrule
\end{tabular}

\medskip
{\textbf{B. Metamorphic-discrepancy change by relation (\%)}}\par\smallskip
\setlength{\tabcolsep}{10pt}
\renewcommand{\arraystretch}{1.06}
\begin{tabular}{@{}lrrrr@{}}
\toprule
Relation & MPNN & LapPE+MPNN & RWSE+MPNN & LUMINA \\
\midrule
\texttt{bus-relabel} & \textemdash & $97.3$ & \textemdash & \textemdash \\
\texttt{branch-reversal} & \textemdash & \textemdash & \textemdash & $74.7$ \\
\texttt{ref-bus-move} & $37.1$ & $36.9$ & $-3.3$ & $21.5$ \\
\texttt{component-split} & $24.9$ & $-15.7$ & $29.9$ & $26.5$ \\
\texttt{parallel-split} & $31.5$ & $96.3$ & $20.1$ & $28.6$ \\
\texttt{island-union} & \textemdash & $96.4$ & \textemdash & \textemdash \\
\texttt{basekv-scale} & \textemdash & \textemdash & \textemdash & $92.5$ \\
\texttt{tap-side-flip} & $45.4$ & $49.9$ & $-21.2$ & $75.3$ \\
\texttt{load-as-gen} & $36.1$ & $-13.4$ & $50.0$ & $75.9$ \\
\texttt{null-pad} & $44.4$ & $-29.4$ & $15.6$ & $60.4$ \\
\texttt{island-ref-mix} & $37.0$ & $96.4$ & $-2.9$ & $21.7$ \\
\bottomrule
\end{tabular}
\par
}
}
\end{table}

Panel A of \cref{tab:aug-appendix} takes the median over the relations used to train the models. Per-channel prediction MAEs appear in \cref{tab:augmentation-summary}.

\subsection{Accuracy and relation-level effects}
\label{app:augmentation-accuracy}

On transformed scenarios, mean MAE improves by $+87.0\%$ for LapPE+MPNN, $+41.9\%$ for LUMINA, and $+1.1\%$ for RWSE+MPNN. MPNN has an improvement of $-3.2\%$, meaning its error increases. Population standard deviations across the seed-level changes are $1.2$, $1.5$, $0.7$, and $3.3$ percentage points, respectively. \crrev{Most of the LapPE+MPNN gain reflects an eigensolver mismatch between training and this audit (\cref{app:cohort-provenance}).} LapPE+MPNN, RWSE+MPNN, and LUMINA improve in all three seeds; MPNN improves in one of three. \crrev{On canonical scenarios, improvements are $+86.8\%$ for LapPE+MPNN, $+0.4\%$ for RWSE+MPNN, $-3.9\%$ for MPNN, and $-3.0\%$ for LUMINA.} MPNN and LUMINA worsen in every seed.

Median normalized discrepancy $\Delta_r$ falls by $30.1\%$, $16.2\%$, $36.1\%$, and $70.8\%$ for MPNN, LapPE+MPNN, RWSE+MPNN, and LUMINA, respectively. It decreases in 11 of 12 model--seed pairs; one LapPE+MPNN seed is almost unchanged at $-0.1\%$. Results differ by relation. \crrev{LapPE+MPNN reduces discrepancy by $96$ to $97\%$ on its four spectral failures: \texttt{bus-relabel}, \texttt{island-union}, \texttt{island-ref-mix}, and \texttt{parallel-split}. These relations expose the arbitrary signs, bases, and ordering of its raw Laplacian eigenvectors.}

\crrev{These gains coexist with regressions. LapPE+MPNN worsens on \texttt{null-pad} ($-29.4\%$ improvement), \texttt{component-split} ($-15.7\%$), and \texttt{load-as-gen} ($-13.4\%$). RWSE+MPNN worsens on} \texttt{tap-side-flip} ($-21.2\%$), \texttt{ref-bus-move} ($-3.3\%$), and \texttt{island-ref-mix} ($-2.9\%$). \crrev{Reducing selected spectral discrepancies while worsening other relations does not establish a complete repair.} MPNN improves on all seven trained relations and LUMINA on all nine. \crrev{The aggregate median summarizes the overall direction but cannot establish improvement on every relation.}

\begin{crrevision}
\subsection{Feasibility and out-of-distribution behavior}
\label{app:augmentation-feasibility}

No evaluated prediction meets both $10^{-6}$ tolerances. The feasible-source rate is therefore 0\% for every arm: none of the 50 sources per seed has all 53 views feasible. The scenario-clustered equality and pooled inequality P95 values fall with augmentation for all four models, but these reductions do not reach AC feasibility.

Panel A of \cref{tab:penalty-violations-scaled} reports absolute violations for each constraint family separately. Within each seed, a source contributes its worst value over all evaluated views. The table gives the mean $\pm$ sample standard deviation across the seed-level medians, P95 values, and maxima. Active and reactive power, generator bounds, and thermal limits are in per unit on the system's 100-MVA base. Voltage is in per unit and angle-difference excess in radians. Keeping the families separate avoids taking an absolute maximum across incompatible units. The LUMINA baseline violates angle-difference limits even though all other arms have zero violations in that family.

Panel B expresses violations relative to physical scales. In Panel B1, the nodal complex-power residual is $\sqrt{r_{P,b}^2+r_{Q,b}^2}$. Its maximum is divided by the source scenario's total apparent demand, $\sum_b\sqrt{P_{d,b}^2+Q_{d,b}^2}$, using the same denominator for every transformed view of that source. The canonical metric averages these values within each seed; improvement is calculated within seed before averaging across seeds. In Panel B2, each positive inequality excess is divided by the relevant element's nominal voltage, generator capacity range, branch rating, or violated angle limit. We then take the worst view per source before computing the P95 and maximum. These percentages give scale to each family's violations; absolute tolerances still determine feasibility. All evaluated denominators were present, finite, nonzero, and active, so no element was excluded from relative aggregation.

\begin{table}[p]
\crcolor
\caption{AC constraint violations from the augmentation training experiment. Panel A reports absolute family-level violations as baseline $\to$ augmented; each side is the mean $\pm$ sample standard deviation across three seed-level statistics after clustering 53 views under each of 50 source scenarios. P95 uses linear interpolation. Panel B reports normalized context using scenario- or element-specific denominators. The AC-feasible source-scenario rate is 0\% for every arm at the $10^{-6}$ tolerance.}
\label{tab:penalty-violations-scaled}
\centering
\scriptsize
{\textbf{A. Absolute violations by constraint family}}\par\smallskip
\setlength{\tabcolsep}{2.5pt}
\renewcommand{\arraystretch}{0.92}
\resizebox{\linewidth}{!}{%
\begin{tabular}{@{}lllrrrrrr@{}}
\toprule
\multirow{2}{*}{Model} & \multirow{2}{*}{Constraint family} & \multirow{2}{*}{Unit} & \multicolumn{2}{c}{Median} & \multicolumn{2}{c}{P95} & \multicolumn{2}{c}{Maximum} \\
\cmidrule(lr){4-5}\cmidrule(lr){6-7}\cmidrule(l){8-9}
 & & & Base & Aug. & Base & Aug. & Base & Aug. \\
\midrule
\multirow{7}{*}{MPNN} & Active-power balance & p.u. & $21.1 \pm 2.39$ & $13.2 \pm 1.59$ & $58.0 \pm 18.7$ & $42.2 \pm 2.26$ & $64.1 \pm 16.5$ & $50.8 \pm 6.4$ \\
 & Reactive-power balance & p.u. & $8.47 \pm 0.476$ & $6.5 \pm 0.746$ & $24.2 \pm 9.59$ & $14.9 \pm 1.68$ & $36.9 \pm 10.2$ & $22.3 \pm 1.14$ \\
 & Voltage bounds & p.u. & $0.00756 \pm 0.00303$ & $0.00617 \pm 0.00361$ & $0.0288 \pm 0.0149$ & $0.0225 \pm 0.0133$ & $0.0406 \pm 0.0186$ & $0.0491 \pm 0.0351$ \\
 & Generator active-power bounds & p.u. & $0.206 \pm 0.116$ & $0.0976 \pm 0.0385$ & $2.68 \pm 0.482$ & $0.647 \pm 0.215$ & $10.2 \pm 2.16$ & $1.29 \pm 0.468$ \\
 & Generator reactive-power bounds & p.u. & $0.0333 \pm 0.00659$ & $0.0272 \pm 0.00723$ & $0.138 \pm 0.041$ & $0.109 \pm 0.0848$ & $0.418 \pm 0.173$ & $0.163 \pm 0.0549$ \\
 & Branch thermal limits & p.u. & $4.92 \pm 1.82$ & $2.5 \pm 0.253$ & $19.3 \pm 4.16$ & $11.9 \pm 1.57$ & $28.8 \pm 11.4$ & $20.2 \pm 3.97$ \\
 & Angle-difference limits & rad & $0 \pm 0$ & $0 \pm 0$ & $0 \pm 0$ & $0 \pm 0$ & $0 \pm 0$ & $0 \pm 0$ \\
\addlinespace
\multirow{7}{*}{LapPE+MPNN} & Active-power balance & p.u. & $89.2 \pm 29.3$ & $12.6 \pm 2.86$ & $114 \pm 42.1$ & $39.5 \pm 5.17$ & $128 \pm 45.1$ & $52.0 \pm 17.5$ \\
 & Reactive-power balance & p.u. & $38.2 \pm 5.89$ & $6.18 \pm 1.14$ & $51.3 \pm 2.59$ & $12.2 \pm 1.76$ & $61.5 \pm 6.8$ & $14.7 \pm 2$ \\
 & Voltage bounds & p.u. & $0.0447 \pm 0.0123$ & $0.00566 \pm 6.30\times10^{-4}$ & $0.0583 \pm 0.0147$ & $0.0162 \pm 0.0048$ & $0.0777 \pm 0.0302$ & $0.0464 \pm 0.0117$ \\
 & Generator active-power bounds & p.u. & $1.32 \pm 0.712$ & $0.0828 \pm 0.0272$ & $2.29 \pm 0.733$ & $0.565 \pm 0.0711$ & $9.16 \pm 4.15$ & $1.04 \pm 0.208$ \\
 & Generator reactive-power bounds & p.u. & $0.23 \pm 0.0212$ & $0.0252 \pm 0.0166$ & $0.404 \pm 0.0964$ & $0.0709 \pm 0.014$ & $0.581 \pm 0.186$ & $0.103 \pm 0.0321$ \\
 & Branch thermal limits & p.u. & $32.9 \pm 5.78$ & $2.48 \pm 0.685$ & $42.1 \pm 4.6$ & $12.7 \pm 3.59$ & $53.1 \pm 5.37$ & $17.9 \pm 7.29$ \\
 & Angle-difference limits & rad & $0 \pm 0$ & $0 \pm 0$ & $0 \pm 0$ & $0 \pm 0$ & $0 \pm 0$ & $0 \pm 0$ \\
\addlinespace
\multirow{7}{*}{RWSE+MPNN} & Active-power balance & p.u. & $19.2 \pm 3.05$ & $10.6 \pm 2.18$ & $51.8 \pm 10.9$ & $33.5 \pm 2.38$ & $74.2 \pm 13.3$ & $46.6 \pm 4.53$ \\
 & Reactive-power balance & p.u. & $8.39 \pm 1.49$ & $5.18 \pm 0.414$ & $19.3 \pm 4.62$ & $12.1 \pm 0.743$ & $30.2 \pm 5.67$ & $13.5 \pm 1.31$ \\
 & Voltage bounds & p.u. & $0.00707 \pm 0.00302$ & $0.00629 \pm 0.00287$ & $0.0206 \pm 0.00408$ & $0.0196 \pm 0.00329$ & $0.0347 \pm 0.00747$ & $0.0333 \pm 0.012$ \\
 & Generator active-power bounds & p.u. & $0.162 \pm 0.0675$ & $0.0709 \pm 0.0325$ & $2.21 \pm 0.666$ & $0.525 \pm 0.112$ & $10.6 \pm 1.41$ & $1.02 \pm 0.211$ \\
 & Generator reactive-power bounds & p.u. & $0.0268 \pm 0.00296$ & $0.0219 \pm 0.00326$ & $0.186 \pm 0.128$ & $0.0868 \pm 0.0609$ & $0.289 \pm 0.103$ & $0.115 \pm 0.0765$ \\
 & Branch thermal limits & p.u. & $4.33 \pm 0.889$ & $2.72 \pm 0.468$ & $18.8 \pm 2.82$ & $10.2 \pm 0.843$ & $25.8 \pm 3.6$ & $13.6 \pm 3.63$ \\
 & Angle-difference limits & rad & $0 \pm 0$ & $0 \pm 0$ & $0 \pm 0$ & $0 \pm 0$ & $0 \pm 0$ & $0 \pm 0$ \\
\addlinespace
\multirow{7}{*}{LUMINA} & Active-power balance & p.u. & $736 \pm 411$ & $91.2 \pm 9.83$ & $871 \pm 342$ & $164 \pm 10.6$ & $989 \pm 401$ & $188 \pm 8.73$ \\
 & Reactive-power balance & p.u. & $568 \pm 512$ & $24.2 \pm 2.75$ & $693 \pm 405$ & $40.3 \pm 5.05$ & $1473 \pm 1007$ & $54.5 \pm 2.02$ \\
 & Voltage bounds & p.u. & $2.38\times10^{-8} \pm 0$ & $0 \pm 0$ & $2.38\times10^{-8} \pm 0$ & $0 \pm 0$ & $2.38\times10^{-8} \pm 0$ & $0 \pm 0$ \\
 & Generator active-power bounds & p.u. & $7.12\times10^{-8} \pm 4.58\times10^{-8}$ & $3.21\times10^{-8} \pm 1.71\times10^{-8}$ & $1.06\times10^{-7} \pm 3.03\times10^{-8}$ & $3.21\times10^{-8} \pm 1.71\times10^{-8}$ & $1.24\times10^{-7} \pm 0$ & $4.20\times10^{-8} \pm 0$ \\
 & Generator reactive-power bounds & p.u. & $6.44\times10^{-9} \pm 7.46\times10^{-9}$ & $0 \pm 0$ & $1.86\times10^{-8} \pm 2.83\times10^{-8}$ & $0 \pm 0$ & $2.86\times10^{-8} \pm 4.56\times10^{-8}$ & $0 \pm 0$ \\
 & Branch thermal limits & p.u. & $238 \pm 119$ & $10.8 \pm 0.727$ & $276 \pm 91.0$ & $20.1 \pm 2.13$ & $502 \pm 386$ & $32.0 \pm 3.52$ \\
 & Angle-difference limits & rad & $10.9 \pm 1.85$ & $0 \pm 0$ & $11.9 \pm 0.445$ & $0 \pm 0$ & $13.2 \pm 1.9$ & $0 \pm 0$ \\
\bottomrule
\end{tabular}%
}
\end{table}

\begin{table}[p]
\crcolor
\centering
{\small\textbf{Table \thetable{} (continued).} AC constraint violations from the augmentation training experiment. Panel B1 gives system-scale equality context and the voltage global maximum. Panel B2 reports relative inequality overruns by family. Each entry is the mean $\pm$ sample standard deviation across three seeds.\par}
\medskip
\scriptsize
{\textbf{B. Normalized interpretability metrics}}\par\smallskip
{\textbf{B1. System-scale equality and voltage context}}\par\smallskip
\setlength{\tabcolsep}{2.5pt}
\renewcommand{\arraystretch}{1.0}
\resizebox{\linewidth}{!}{%
\begin{tabular}{@{}lrrrrrrr@{}}
\toprule
 & \multicolumn{2}{c}{\shortstack{Global-max nodal complex-power balance residual\\(\% total apparent demand)}} & \multicolumn{2}{c}{\shortstack{Global-max voltage-bound excess\\(\% nominal)}} & \multicolumn{3}{c}{\shortstack{Canonical mean maximum nodal complex-power balance residual\\(\% total apparent demand)}} \\
\cmidrule(lr){2-3}\cmidrule(lr){4-5}\cmidrule(l){6-8}
Model & Base & Aug. & Base & Aug. & Base & Aug. & Imp. (\%) \\
\midrule
MPNN & $36.9 \pm 7.29$ & $28.8 \pm 2.42$ & $4.06 \pm 1.86$ & $4.91 \pm 3.51$ & $4.32 \pm 0.484$ & $3.94 \pm 0.393$ & $8.3$ \\
LapPE+MPNN & $72.6 \pm 25.1$ & $28.6 \pm 9.52$ & $7.77 \pm 3.02$ & $4.64 \pm 1.17$ & $29.6 \pm 6.21$ & $3.34 \pm 0.385$ & $88.4$ \\
RWSE+MPNN & $40.4 \pm 7.42$ & $25.7 \pm 2.71$ & $3.47 \pm 0.747$ & $3.33 \pm 1.2$ & $3.78 \pm 0.28$ & $3.31 \pm 0.157$ & $12.2$ \\
LUMINA & $893 \pm 499$ & $103 \pm 5.4$ & $2.38\times10^{-6} \pm 0$ & $0 \pm 0$ & $10.0 \pm 0.493$ & $8.85 \pm 0.481$ & $11.8$ \\
\bottomrule
\end{tabular}%
}

\medskip
{\textbf{B2. Relative inequality overruns (\%)}}\par\smallskip
\setlength{\tabcolsep}{3.5pt}
\renewcommand{\arraystretch}{0.94}
\resizebox{\linewidth}{!}{%
\begin{tabular}{@{}llrrrr@{}}
\toprule
\multirow{2}{*}{Model} & \multirow{2}{*}{Constraint family} & \multicolumn{2}{c}{P95} & \multicolumn{2}{c}{Maximum} \\
\cmidrule(lr){3-4}\cmidrule(l){5-6}
 & & Base & Aug. & Base & Aug. \\
\midrule
\multirow{5}{*}{MPNN} & Voltage bounds & $2.88 \pm 1.49$ & $2.25 \pm 1.33$ & $4.06 \pm 1.86$ & $4.91 \pm 3.51$ \\
 & Generator active-power bounds & $157 \pm 13.4$ & $135 \pm 47.5$ & $270 \pm 77.9$ & $202 \pm 129$ \\
 & Generator reactive-power bounds & $838 \pm 434$ & $628 \pm 282$ & $3212 \pm 2238$ & $2570 \pm 1685$ \\
 & Branch thermal limits & $892 \pm 197$ & $375 \pm 108$ & $1418 \pm 675$ & $708 \pm 158$ \\
 & Angle-difference limits & $0 \pm 0$ & $0 \pm 0$ & $0 \pm 0$ & $0 \pm 0$ \\
\addlinespace
\multirow{5}{*}{LapPE+MPNN} & Voltage bounds & $5.83 \pm 1.47$ & $1.62 \pm 0.480$ & $7.77 \pm 3.02$ & $4.64 \pm 1.17$ \\
 & Generator active-power bounds & $616 \pm 122$ & $165 \pm 104$ & $942 \pm 431$ & $409 \pm 293$ \\
 & Generator reactive-power bounds & $1203 \pm 408$ & $494 \pm 249$ & $2687 \pm 1645$ & $1934 \pm 198$ \\
 & Branch thermal limits & $2098 \pm 288$ & $541 \pm 222$ & $2309 \pm 306$ & $649 \pm 185$ \\
 & Angle-difference limits & $0 \pm 0$ & $0 \pm 0$ & $0 \pm 0$ & $0 \pm 0$ \\
\addlinespace
\multirow{5}{*}{RWSE+MPNN} & Voltage bounds & $2.06 \pm 0.408$ & $1.96 \pm 0.329$ & $3.47 \pm 0.747$ & $3.33 \pm 1.20$ \\
 & Generator active-power bounds & $180 \pm 84.6$ & $216 \pm 33.6$ & $499 \pm 405$ & $679 \pm 485$ \\
 & Generator reactive-power bounds & $688 \pm 548$ & $438 \pm 299$ & $1663 \pm 1174$ & $1534 \pm 939$ \\
 & Branch thermal limits & $803 \pm 73.9$ & $458 \pm 79.9$ & $1141 \pm 229$ & $601 \pm 127$ \\
 & Angle-difference limits & $0 \pm 0$ & $0 \pm 0$ & $0 \pm 0$ & $0 \pm 0$ \\
\addlinespace
\multirow{5}{*}{LUMINA} & Voltage bounds & $2.38\times10^{-6} \pm 0$ & $0 \pm 0$ & $2.38\times10^{-6} \pm 0$ & $0 \pm 0$ \\
 & Generator active-power bounds & $1.19\times10^{-5} \pm 1.23\times10^{-6}$ & $4.76\times10^{-6} \pm 2.62\times10^{-6}$ & $1.19\times10^{-5} \pm 1.23\times10^{-6}$ & $4.85\times10^{-6} \pm 2.53\times10^{-6}$ \\
 & Generator reactive-power bounds & $7.01\times10^{-7} \pm 7.38\times10^{-7}$ & $0 \pm 0$ & $8.63\times10^{-7} \pm 9.99\times10^{-7}$ & $0 \pm 0$ \\
 & Branch thermal limits & $9400 \pm 5026$ & $859 \pm 27.6$ & $9975 \pm 5084$ & $1537 \pm 233$ \\
 & Angle-difference limits & $2274 \pm 85.0$ & $0 \pm 0$ & $2530 \pm 363$ & $0 \pm 0$ \\
\bottomrule
\end{tabular}%
}
\end{table}

Cross-grid generalization results are mixed, and all three seeds agree in direction for only a few model--grid combinations (\cref{tab:aug-ood-clean}).
LapPE+MPNN improves on case14 ($21.8\%$) but worsens on case2000 and case4661 ($7.7\%$ and $13.4\%$). RWSE+MPNN improves on case2000 ($3.5\%$). LUMINA improves on case14, case2000, and case4661 ($11.0\%$, $5.7\%$, and $8.0\%$) and worsens slightly on the in-distribution case500 ($3.4\%$).
LUMINA is the only model with improvements on three unseen grids that each agree in sign across all three seeds. This agreement does not extend to case118, where its mean error increases by $14.2\%$ and seedwise signs are mixed. Every model--grid combination without an asterisk in \cref{tab:aug-ood-clean} has mixed seedwise signs.
Our experiment therefore does not support a general cross-grid benefit from using transformations as data augmentation.
\end{crrevision}

\begin{table}[htbp]
\crcolor
\caption{Effect of transformation augmentation on other grids. \crrev{Entries are average per-seed percent changes in equal-channel MAE, and \textbf{negative} values denote improvement under augmentation.} An asterisk marks entries where all three seeds agree in sign. Case500 is a separate 200-source clean cohort with no scenario overlap with the main 50-source audit; evaluation paths also differ (\cref{app:cohort-provenance}). The other grids receive no additional training.}
\label{tab:aug-ood-clean}
\centering
\small
{
\begin{tabular}{@{}lrrrr@{}}
\toprule
Grid & MPNN & LapPE+MPNN & RWSE+MPNN & LUMINA \\
\midrule
case14 & $-3.8$ & $-21.8^{*}$ & $-4.3$ & $-11.0^{*}$ \\
case118 & $+11.3$ & $-2.8$ & $+13.5$ & $+14.2$ \\
case500 & $+4.8$ & $-0.4$ & $+0.4$ & $+3.4^{*}$ \\
case2000 & $-3.9$ & $+7.7^{*}$ & $-3.5^{*}$ & $-5.7^{*}$ \\
case4661 & $+0.2$ & $+13.4^{*}$ & $+6.0$ & $-8.0^{*}$ \\
\bottomrule
\end{tabular}
}
\end{table}

\begin{crrevision}
\subsection{Cohort and evaluation-path provenance}
\label{app:cohort-provenance}
The main accuracy table and the case500 row of \cref{tab:aug-ood-clean} use different evaluation cohorts and answer different empirical questions. The main table uses the 50-source canonical subset of the transformed-view audit. The cross-grid table uses a separately selected 200-source clean cohort, with no shared case500 scenario IDs. Reaggregating the saved native-unit MAEs reproduces both tables. The cross-grid run records 24 checkpoint hashes and 24,000 model--scenario rows: four models, two arms, three seeds, five grids, and 200 scenarios. Its MPNN-family evaluation is batched; the transformed-view audit uses a different prediction path.

The LapPE baseline has mean canonical MAE $0.09983$ in the main cohort and $0.01253$ in the cross-grid case500 cohort. The corresponding augmented MAEs are $0.01309$ and $0.01246$. A matched follow-up traces the gap mainly to the eigensolver backend. LapPE+MPNN training and cross-grid evaluation use batched single-precision CUDA \texttt{torch.linalg.eigh} to compute Laplacian eigenvectors. The transformed-view audit computes them separately for each scenario using double-precision NumPy (LAPACK) \texttt{eigh} on CPU. Neither path fixes eigenvector signs or bases within repeated eigenspaces.

We re-evaluated the same six checkpoints, with hashes matching the cross-grid run, on the same 200 cross-grid case500 scenarios after changing only the eigensolver backend to CPU. Baseline mean MAE rises from $0.01253$ to $0.08853$ ($7.1\times$; seed values $0.0797$, $0.0992$, and $0.0867$), close to the main-cohort value. Augmented MAE changes from $0.01246$ to $0.01283$ ($+3.0\%$).

Most of the large main-cohort LapPE gain therefore measures augmentation's robustness to CPU/LAPACK spectral coordinates after CUDA training. It does not demonstrate an accuracy gain under the training backend or a general repair of spectral encodings. LapPE+MPNN's main-audit baseline violation P95 and normalized discrepancies use the same CPU path and may also be inflated by the mismatch; this follow-up does not quantify those effects. The remaining cross-grid values describe transfer under their stated cohorts and sign convention.
\end{crrevision}

\clearpage
\section{Design and transparency implications}
\label{app:discussion}

\subsection{Designing consistency by construction}
\label{app:exact-construction}

An admitted relation links two encodings of the same dispatch problem, with predictions compared after the declared output map. \crrev{A material difference between those aligned predictions means that the deterministic surrogate depends on the encoding as well as on the physical problem.} Low reference-label error on one serialization can miss this dependence.

\crrev{A relation enforced by construction holds for any learned weights, whether or not it appeared in training.} The architecture screen points to positional summaries that preserve permutation equivariance, output partitions based on component type or electrical role, and symmetric transport of orientation-dependent branch attributes. \crrev{Preprocessing can address some failures by standardizing units, angle references, and row conventions.} Relations that change component count or type instead require common component encodings and additive aggregation at physical buses or devices. \crrev{These changes remain compatible with graph models, provided each proposed remedy addresses the relation that failed.}

\crrev{Where an exact construction is unavailable, training-time augmentation can reduce sensitivity, including for \ELC{} relations that change graph size. Its effect is approximate and depends on optimization. In our experiment, aggregate representation sensitivity falls, but some relations worsen, canonical error can increase or decrease, and every prediction remains infeasible.} A useful AC-OPF surrogate therefore needs exact consistency for representation and angle-reference choices where practical, measured consistency for the remaining equivalences, and separate tests of per-channel accuracy, AC feasibility, bounds, and dispatch cost.

\subsection{Minimum transparency package}
\label{app:transparency-package}

To reproduce a consistency verdict, an independent auditor must construct both views of a case and align their outputs. The following information makes those steps possible. \crrev{A versioned hosted endpoint can substitute for released parameters if it exposes the same documented inputs and outputs.}

\begin{itemize}
\item \crrev{Runnable inference requires a versioned checkpoint with checksums and a pinned environment, or a versioned endpoint that accepts documented case files.}
\item \crrev{The input conventions cover schema and feature order, physical units and per-unit bases, the angle-reference convention, and normalization statistics.}
\item \crrev{The output schema gives the physical quantity, unit, and stable bus or device identity of each output channel.}
\item \crrev{The training topologies list the grids or grid families represented in training.}
\item \crrev{Each claimed invariance names a relation family that preprocessing, architecture, or training is said to enforce.}
\item \crrev{A worked example pairs one case file with its encoded model input and numerical canonical outputs.}
\end{itemize}

\clearpage
\end{document}